\documentclass[journal]{IEEEtran}
\usepackage{enumerate}
\usepackage{amssymb}
\usepackage{amsmath}

\usepackage{graphicx}
\usepackage{subfigure}
\usepackage{bm}
\usepackage{cite}
\usepackage{multirow}
\usepackage{color}

\usepackage{makecell}
\usepackage[table,xcdraw]{xcolor}
\usepackage{booktabs} 
\usepackage{graphicx}
\usepackage[colorlinks=true, linkcolor=blue, urlcolor=blue, citecolor=blue]{hyperref}
\usepackage{algorithm}
\usepackage{algpseudocode}
\usepackage{diagbox}

\ifCLASSINFOpdf
\else
\fi
\begin{document}
\title{Non-target Divergence Hypothesis: Toward Understanding Domain Gaps in Cross-Modal Knowledge Distillation}
\author{Yilong~Chen,
        Zongyi~Xu,
        Xiaoshui~Huang,
        Shanshan~Zhao,
        Xinqi~Jiang,
        Xinyu~Gao,
        and~Xinbo~Gao,~\IEEEmembership{Fellow,~IEEE}}
\markboth{IEEE TRANSACTIONS ON MULTIMEDIA}
{Shell \MakeLowercase{\textit{et al.}}: Bare Demo of IEEEtran.cls for IEEE Journals}



\maketitle

\begin{abstract}
Compared to single-modal knowledge distillation, cross-modal knowledge distillation faces more severe challenges due to domain gaps between modalities. Although various methods have proposed various solutions to overcome these challenges, there is still limited research on how domain gaps affect cross-modal knowledge distillation. This paper provides an in-depth analysis and evaluation of this issue. We first introduce the Non-Target Divergence Hypothesis (NTDH) to reveal the impact of domain gaps on cross-modal knowledge distillation. Our key finding is that domain gaps between modalities lead to distribution differences in non-target classes, and the smaller these differences, the better the performance of cross-modal knowledge distillation. Subsequently, based on Vapnik-Chervonenkis (VC) theory, we derive the upper and lower bounds of the approximation error for cross-modal knowledge distillation, thereby theoretically validating the NTDH. Finally, experiments on five cross-modal datasets further confirm the validity, generalisability, and applicability of the NTDH.

\end{abstract}


% \fi
\IEEEpeerreviewmaketitle

\section{Introduction}

\IEEEPARstart{I}{n} recent years, cross-modal knowledge distillation (KD) has expanded the traditional KD approach to encompass multimodal learning, achieving notable success in various applications \cite{wang2021knowledge,liu2023dccd,liu2024fine,singh2024kl}. However, when there are considerable domain gaps in cross-modal KD, even a more accurate teacher model may not effectively instruct the student model. To overcome these challenges, many researchers have sought to enhance the effectiveness of cross-modal KD by designing efficient fusion strategies \cite{sun2024uni,liu2023new,tan2022improving,ding2022dual} or developing novel loss functions \cite{romero2014fitnets,tian2019contrastive,park2019relational,passalis2018probabilistic,tung2019similarity,zhao2022decoupled}. However, most of these methods focus on complex technical designs with less emphasis on exploring their theoretical foundations, which is the focus of this paper.

\par
\begin{figure}[!t]
\centering
\includegraphics[width=0.95\linewidth]{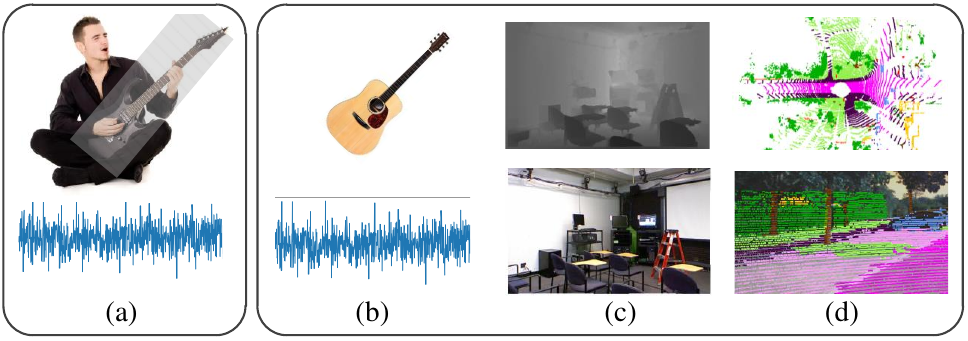}
\caption{Some cross-modal data instances. (a) Modal misaligned scene. (b)-(d) Modal alignment scenario (Our research subjects), such as images and audio of a guitar, RGB images from the same camera perspective and point cloud projected onto images and depth maps.}
\label{fig:1}
\end{figure}

 Xue et al. \cite{xue2023modality} are the first to theoretically focus on KD under modality differences, proposing the Modality Focus Hypothesis (MFH). This hypothesis posits that the performance of cross-modal KD hinges on the modality-general decisive features preserved in the teacher network. These features indicate the degree of alignment between different modalities, with greater alignment leading to improved KD outcomes. Fig. \ref{fig:1}(a) shows an example of a multimodal dataset with both audio and images, where the image data includes not only the scene of guitar music but also background information, leading to incomplete modality alignment; according to the MFH, improved feature alignment is expected to enhance cross-modal KD effectiveness. However, the above hypothesis has two shortcomings: \textit{(1) It cannot explain why cross-modal KD might still fail even when there is modality alignment, as shown in situations like those in Fig. \ref{fig:1}(b)-(d). (2) It lacks a mathematical definition of modality-general key features, making it difficult to identify these features in practical applications.}

\par To address these deficiencies, we first pre-register or align the modalities involved in our study to eliminate the effects caused by unregistered modalities, thereby focusing our research on domain gaps. We adopt this approach because we believe that misalignment is merely an external manifestation, while the actual domain gaps between modalities are the core distinction between cross-modal and single-modal KD.

Secondly, we clearly define the key factors affecting cross-modal KD. Specifically, inspired by \cite{li2023decoupled}, we divide classification predictions into two levels: \textbf{(1) Target class prediction distribution:} a binary prediction distribution for the target category and all non-target categories; \textbf{(2) Non-target class prediction distribution:} a multi-category prediction distribution for each non-target category. We find that for cross-modal KD, the distribution divergence in non-target categories is the decisive factor. The smaller the divergences, the better the effect of cross-modal KD, which we refer to as the Non-target Divergence Hypothesis. Since the distance of non-target distributions can be easily measured using existing distance functions, they can be explicitly defined and calculated. For example, in multimodal point cloud semantic segmentation, both point clouds and images classify road areas. These predictions can be divided into road class prediction distribution and non-road class prediction distribution. If the distribution difference of the non-road class predictions is large, the effectiveness of cross-modal KD is significantly affected and may even fail, as shown in Fig. \ref{fig:2}. The purpose of this work is to prove the validity of the hypothesis from a theoretical analysis perspective. Our major contributions are the following:

\begin{figure}[!t]
\centering
\includegraphics[width=0.95\linewidth]{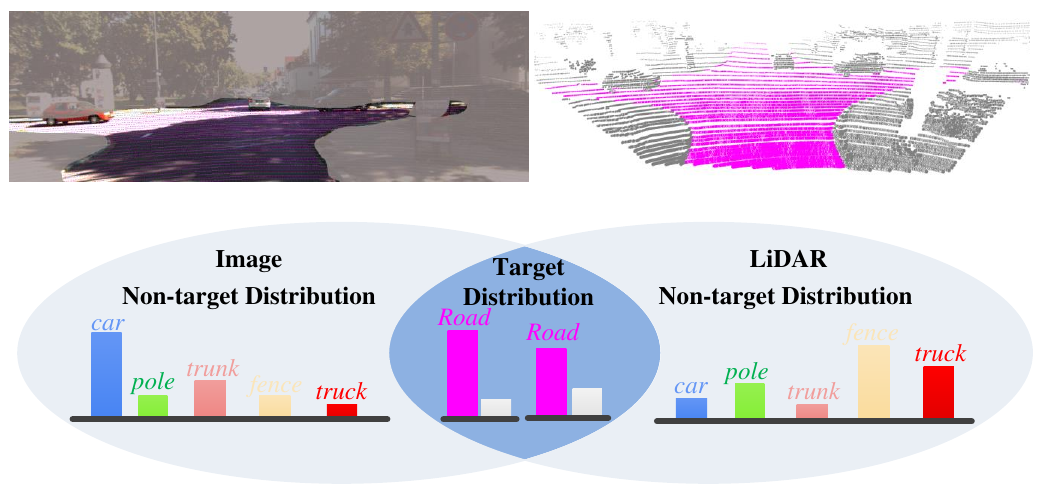}
\caption{Venn diagram example for NTDH. The prediction distributions of the teacher and student models consist of target class and non-target class distributions. For example, in the prediction of road areas, the overlapping region in the Venn diagram represents the target class prediction distribution, while the non-overlapping region represents the non-target class prediction distribution.}
\label{fig:2}
\end{figure}

\begin{itemize}
  \item [1)]
  We propose the Non-target Divergence Hypothesis (NTDH), which posits that the divergence of non-target distributions between modalities is a key factor determining the effectiveness of cross-modal KD.
  \item [2)]
  We theoretically prove the upper and lower bounds of the cross-modal KD approximation error, thereby validating the rationality of NTDH.
  \item [3)]
  We design a weight regulation method and a masking method to experimentally verify the hypothesis. Experimental results on five multimodal datasets support the proposed NTDH.
 
\end{itemize}

\section{Related Work}\label{sec:2}

\subsection{Unimodal KD}

\par KD is a general technique for transferring knowledge from a teacher network to a student network, widely applied in various vision tasks \cite{wang2021knowledge,liu2023dccd,liu2024fine,singh2024kl}. Although progress has been made in improving distillation techniques and exploring new application areas, research on the mechanisms underlying KD remains limited \cite{phuong2019towards,cho2019efficacy,tang2020understanding,ren2022co}. For example, Cho et al. \cite{cho2019efficacy} and Mirzadeh et al. \cite{mirzadeh2020improved} investigate KD in the context of model compression, focusing on the performance of student and teacher networks when their model sizes differ. They point out that a mismatch in capacity between the student and teacher networks can lead to KD failure. Ren et al. \cite{ren2022co} analyze KD in vision transformers and find that the inductive bias of the teacher model is more important than its accuracy in improving the performance of the student model. However, the aforementioned methods mainly focus on the factors affecting unimodal KD, while research on cross-modal KD remains limited.

\subsection{Crossmodal KD}

With the widespread use of the internet and the increasing application of multimodal sensors, there is a surge of interest in multimodal learning \cite{gou2021knowledge}. In line with this trend, KD extends to cross-modal KD, which commonly applies in three scenarios. First, \textbf{Knowledge Expansion}: Cross-modal KD compensates for insufficient data in a single modality. For instance, Ahmad et al. \cite{ahmad2024multi} use multimodal MRI data to overcome clinical data scarcity, while Li et al. \cite{li2023efficient} transfer RGB-trained models to infrared data, addressing its data limitations. Similar approaches include using \textit{Galvanic Skin Response} signals to offset \textit{Electroencephalogram} data collection challenges \cite{liu2023emotionkd}, supplementing \textit{Synthetic Aperture Radar} images with RGB data \cite{nair2024let}, and transferring English-trained models to other languages \cite{weng2024zero}. Second, \textbf{Multimodal Knowledge Fusion}: Cross-modal KD fuses complementary information from different modalities into a unimodal network, requiring multimodal data only during training and simplifying deployment \cite{wang2023cross}, such as using Lidar for RGB images \cite{dai2021learning}, RGB for point clouds \cite{chen2022bevdistill}, facial images for voice data \cite{jin2023cross}, and RGB to enhance thermal imaging \cite{feng2023cekd}. Large language models also aid in cross-modal retrieval \cite{li2024ckdh}. Third, \textbf{Additional Constraints}: In some cases, significant differences in network structures for different modalities make feature layer fusion highly challenging. The loss function of cross-modal knowledge distillation then acts as a regularization term, enforcing consistency constraints on outputs to enhance the performance of individual modalities. For instance, Sarkar et al. \cite{sarkar2024xkd} use KL divergences to align audio and video outputs, while another method, MCKD \cite{ma2023using}, applies multimodal contrastive KD for video-text retrieval. Although these methods show potential in cross-modal KD, most widely used approaches still rely on single-modal techniques, raising questions about their effectiveness and limitations. This paper, therefore, analyzes key factors influencing cross-modal KD to enhance its application.

\subsection{Domain Gaps in Cross-modal KD}

\par In cross-modal KD, domain gaps critically impact performance. To reduce these gaps, modality alignment methods are often employed to unify different modalities from various domains into a common \cite{liu2024taming,chen2024vision,jin2023cross,sarkar2024xkd,yun2023dense,li2023decoupled} or intermediate modality \cite{chen2023enhanced}. However, this approach may result in the loss of modality-specific characteristics. To address this, some methods incorporate decoupling strategies to preserve these characteristics, such as using independent detection heads \cite{lee2023decomposed} for KD or employing feature partitioning to effectively transfer knowledge from the teacher modality while preserving unique features of the student modality \cite{kim2024labeldistill}. Additionally, other approaches focus on minimizing domain gaps by filtering out features with significant discrepancies. For example, Zhuang et al. \cite{zhuang2021perception} and Wang et al. \cite{wang2023learnable} focus on evaluating and filtering significant domain discrepancies, while Huo et al. \cite{huo2024c2kd} selectively filter out misaligned samples to avoid modality imbalance. Although these methods have made some progress, they primarily address multimodal feature fusion and are not directly applicable to dual-branch networks in cross-modal KD. To address this issue, this paper proposes a mask-based approach to mitigate the impact of domain shifts on cross-modal KD.

\section{The Proposed Hypothesis} \label{sec:3}
In this section, we first define the symbols and present the necessary assumptions for subsequent proofs. Next, we introduce the Non-target Divergence Hypothesis and provide a detailed explanation of this hypothesis based on experimental results on Scikit-learn dataset \cite{pedregosa2011scikit}. Finally, we prove the hypothesis using Vapnik-Chervonenkis (VC) theory \cite{vapnik1999overview}.

\subsection{Symbol Definitions And Conditional Assumptions} \label{subsec:3.1}

\par We provide a comprehensive overview of the fundamentals of KD and introduce the symbols used in this study. Although our focus is primarily on C-class classification problems, the concepts discussed are also applicable to regression tasks. To maintain the generality of the framework, we consider a two-modal setting, where the data is represented as ${{x}^{a}}$ and ${{x}^{b}}$, corresponding to the data from modalities `A' and `B', respectively, as shown in Eq. (\ref{eq:01}).

\begin{equation}
\left\{ \begin{matrix}
   \{(x_{i}^{a},{{y}_{i}})\}_{i=1}^{n}\sim{{P}^{n}}({{x}^{a}},y),x_{i}^{a}\in {{\mathbb{R}}^{{{d}_{a}}}},{{y}_{i}}\in {{\Delta }^{c}}  
   \vspace{10pt}\\  % 添加垂直空间，单位可以自行调整
   \{(x_{i}^{b},{{y}_{i}})\}_{i=1}^{n}\sim{{P}^{n}}({{x}^{b}},y),x_{i}^{b}\in {{\mathbb{R}}^{{{d}_{b}}}},{{y}_{i}}\in {{\Delta }^{c}}  \\
\end{matrix} \right.,
\label{eq:01}
\end{equation}

where $(x_{i}^{a},{{y}_{i}})$ and $(x_{i}^{b},{{y}_{i}})$ respectively represent the feature-label pair of modality `A' and modality `B', and ${{\Delta }^{c}}$ represents the set of c-dimensional probability vectors. 

Suppose that our goal is to train a student network that takes ${{x}^{b}}$ as input. In the case of cross-modal KD, the teacher network takes the training data ${{x}^{a}}$ as input and minimizes the training objective as follows:

\begin{equation}
{{f}_{t}}=\underset{f\in {{\mathcal{F}}_{t}}}{\mathop{\arg \min }}\,\frac{1}{n}\sum\limits_{i=1}^{n}{\mathcal{L}({{y}_{i}},\sigma (f(x_{i}^{a})))}+\Omega (\left\| f \right\|) ,
\label{eq:02}
\end{equation}

where ${{\mathcal{F}}_{t}}$ is a class of functions from ${{\mathbb{R}}^{{{d}_{a}}}}$ to ${{\mathbb{R}}^{c}}$ , the function $\sigma $ : ${{\mathbb{R}}^{c}}\to {{\Delta }^{c}}$ is the softmax operation

\begin{equation}
\sigma {{(z)}_{k}}=\frac{{{e}^{zk}}}{\sum\nolimits_{j=1}^{c}{{{e}^{{{z}_{j}}}}}}.
\label{eq:03}
\end{equation}

For all $1\le k\le c$ , the function  $\mathcal{L}:{{\Delta }^{c}}\times {{\Delta }^{c}}\to \mathbb{R}$ is Kullback-Leibler(KL) Divergence \cite{kullback1951information}, and $\Omega :\mathbb{R}\to \mathbb{R}$ is an increasing function which serves as a regularizer.

After training the teacher model ${{f}_{t}}$ using the data $x_{i}^{a}$ from modality `A', our goal is to transfer the knowledge acquired by the teacher network to the student network operating in modality `B'. Therefore, in addition to minimizing the KL loss between the student output and one-hot label, it is also required to minimize the KL loss between the teacher and student outputs. The objective of optimizing the student network is as follows:

\begin{equation}
\begin{aligned}
{{f}_{s}} &= \underset{f \in \mathcal{F}_{s}}{\mathop{\arg \min }}\,\frac{1}{n}\sum\limits{_{i=1}^{n}} \big[(1-\lambda) \cdot \mathcal{L}({{y}_{i}},\sigma (f(x_{i}^{b}))) \\
&\quad + \lambda \cdot \mathcal{L}({{s}_{i}},\sigma (f(x_{i}^{b})))\big],
\end{aligned}
\label{eq:04}
\end{equation}

where ${{s}_{i}}=\sigma ({{f}_{t}}(x_{i}^{a})/T)\in {{\Delta }^{c}}$  represents the soft predictions obtained from ${{f}_{t}}$ about the training on modality `A'. The temperature parameter $T$ ($T>0$)  controls the level of softening or smoothing of the class-probability predictions from ${{f}_{t}}$ and the imitation parameter $\lambda \in [0,1]$ determines the balance between imitating the soft predictions ${{s}_{i}}$ and predicting the true hard labels ${{y}_{i}}$.

Given that the primary focus of this paper is the impact of domain gaps in cross-modal KD, we have listed the following assumptions conditions to control variables and exclude the interference of model capacity and modality strength. These assumptions form the foundational conditions for the discussion in this paper and the basis for the theoretical reasoning in Sec. \ref{subsec:3.3}.

\begin{itemize}
 \item  \textbf{Assumption 1:} ${\mathcal{F}_{s}}$ and ${\mathcal{F}_{t}}$  have the same model capacity, meaning they have the same ability to fit or learn complexity or accommodate information.

\item \textbf{Assumption 2:} $x_{i}^{a}$  and $x_{i}^{b}$ have the same modality strength, meaning that when the same model is trained using $x_{i}^{a}$  and  $x_{i}^{b}$ as data separately, the difference in model prediction accuracy is not significant.

\end{itemize}

\subsection{Non-target Divergence Hypothesis}
\label{subsec:3.2}

\begin{figure*}[!t]
\centering
\includegraphics[width=0.95\linewidth]{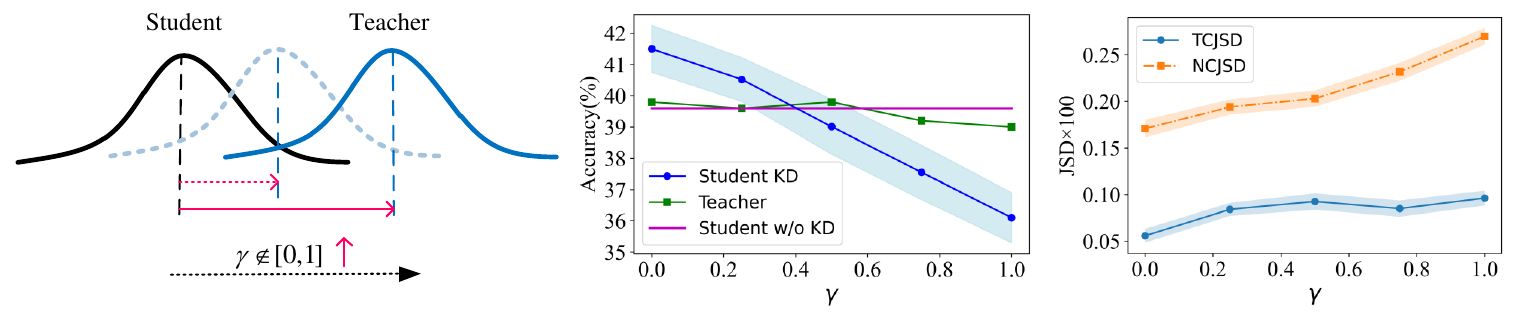}
\caption{An illustration of NTDH with synthetic Scikit-learn data. As the domain discrepancy of multimodal data increases ($\gamma$ ranges from 0 to 1), the performance of KD gradually declines. At the same time, the discrepancy in non-target class prediction distributions increases, and its growth rate far exceeds that of the target class distribution discrepancy.}
\label{fig:3}
\end{figure*}

\par Under the assumption in Sec. \ref{subsec:3.1} and based on VC theory \cite{vapnik1999overview}, the upper and lower bounds of the cross-modal KD approximation error have been derived (see Sec. \ref{subsec:3.3} for the omitted proof). According to this conclusion, it can be inferred that the divergence in non-target class prediction distributions between the teacher and student networks is a key factor affecting the effectiveness of cross-modal KD. This divergence stems from inherent domain discrepancies between modalities, which hinder effective guidance from the teacher to the student. This leads to  the non-target divergence hypothesis.

\textbf{Non-target Divergence Hypothesis (NTDH)}: \textit{For cross-modal KD, the performance of KD is determined by the distribution divergence of non-target classes. The smaller this divergence, the better the student network performs.}

This hypothesis posits that cross-modal KD benefits from the consistency in the distribution of non-target classes. Furthermore, it explains the observation that, in certain circumstances, the performance of the teacher network is not directly correlated with the performance of the student network.

To intuitively understand our hypothesis, we use the Scikit-learn toolkit to simulate six sets of multimodal data, where $\gamma \in \left[ 0,1 \right]$ represents the degree of domain discrepancy between modalities, with higher values indicating greater domain differences, as shown in Fig. \ref{fig:3}. To satisfy Assumptions 1 and 2, both the teacher and student networks use the same architecture and maintain consistency in data intensity (see Sec. \ref{subsec:3.3} for details). We conduct experiments on these six datasets and record the Jensen-Shannon Divergence between the teacher and student networks in both target and non-target distributions during stable training, denoted as TCJSD and NCJSD, respectively:

\begin{equation}
\begin{aligned}
\text{TCJSD} =\frac{1}{2}\cdot p_{k}^{ta}\log \left( \frac{p_{k}^{ta}}{{{m}_{k}}} \right)+\frac{1}{2}\cdot p_{k}^{sb}\log \left( \frac{p_{k}^{sb}}{{{m}_{k}}} \right),
\end{aligned}
\label{eq:05}
\end{equation}

\begin{equation}
\begin{aligned}
\text{NCJSD} = \sum\limits_{i=1,i\ne k}^{C} \left( \frac{1}{2} \cdot \hat{p}_{i}^{ta} \log \left( \frac{\hat{p}_{i}^{ta}}{{m}_{i}} \right) + \frac{1}{2} \cdot \hat{p}_{i}^{sb} \log \left( \frac{\hat{p}_{i}^{sb}}{{m}_{i}} \right) \right),
\end{aligned}
\label{eq:06}
\end{equation}

where ${{m}_{i}}=\frac{1}{2}\cdot \left( \hat{p}_{i}^{ta}+\hat{p}_{i}^{sb} \right)$,
${{m}_{k}}=\frac{1}{2}\cdot \left( p_{k}^{ta}+p_{k}^{sb} \right)$,
\\
\[
\begin{array}{rl}
p_{k}^{ta} & = \sigma ({{f}_{t}}(x_{k}^{a})) = \frac{\exp ({{f}_{t}}(x_{k}^{a}))}{\sum\limits_{j=1}^{C}{\exp ({{f}_{t}}(x_{j}^{a}))}}, \\
p_{k}^{sb} & = \sigma ({{f}_{s}}(x_{k}^{b})) = \frac{\exp ({{f}_{s}}(x_{k}^{b}))}{\sum\limits_{j=1}^{C}{\exp ({{f}_{s}}(x_{j}^{b}))}}, \\
\hat{p}_{i}^{ta} & = \sigma ({{f}_{t}}(\hat{x}_{i}^{a})) = \frac{\exp ({{f}_{t}}(\hat{x}_{i}^{a}))}{\sum\limits_{j=i, j\ne k}^{C}{\exp ({{f}_{t}}(x_{j}^{a}))}}, \\
\hat{p}_{i}^{sb} & = \sigma ({{f}_{s}}(\hat{x}_{i}^{b})) = \frac{\exp ({{f}_{s}}(\hat{x}_{i}^{b}))}{\sum\limits_{j=i, j\ne k}^{C}{\exp ({{f}_{s}}(x_{j}^{b}))}}.
\end{array}
\]

The experimental results indicate that as the domain discrepancy between modalities increases (i.e., as $\gamma$ increases from 0 to 1), the accuracy of the teacher network remains stable, while the performance of the student network declines significantly. Concurrently, there is a significant increase in the distance of the non-target distribution, whereas the distance of the target distribution remains almost unchanged, as illustrated in Fig. \ref{fig:3}. These findings suggest that in cross-modal KD, the greater the distribution divergence of non-target classes, the lower the effectiveness of KD. In contrast, a smaller discrepancy leads to better distillation outcomes. This is a qualitative observation; further validation of these conclusions will be provided through theoretical derivations in Sec. \ref{subsec:3.3}.

\subsection{Prove the Non-target Divergence Hypothesis}
\label{subsec:3.3}

% \begin{table}[htb]
% \centering
% % \caption{}
% \label{tab:symbol}
% \begin{tabular}{c@{\hspace{2cm}}l}
% % \begin{tabular}{ll}
% \hline
% \multicolumn{2}{c}{Symbol List}        \\ \hline
% ${{x}^{a}}$    & data   from modalities ‘A’       \\
% ${{x}^{b}}$    & data   from modalities ‘B’        \\
% ${\mathcal{F}_{t}}$    & teacher   function  (trained on ${{x}^{a}}$)               \\
% ${\mathcal{F}_{s}}$    & student   function (trained on ${{x}^{a}}$ or ${{x}^{b}}$)  \\
% $T$    & temperature parameter                 \\
% ${{y}_{i}}$    & the true hard labels                  \\
% ${{s}_{i}}$    & soft predictions                      \\
% ${\varepsilon}$   & approximation error                  \\
% $O(\cdot )$ & Estimation   error                      \\
% $\sigma $   & the softmax operation.  \\
% ${{\left| \cdot  \right|}_{C}}$ & somes function class capacity measure\\
% $n$   & The number of data points       \\ \hline
% \end{tabular}
% \end{table}

\par Recall our three actors: the student function ${{f}_{s}}\in {{\mathcal{F}}_{s}}$  (trained on $x_{i}^{b}$ ), the teacher function ${{f}_{t}}\in {{\mathcal{F}}_{t}}$ (trained on $x_{i}^{a}$ or $x_{i}^{b}$), and the real target function of interest to both the student and the teacher, $f\in \mathcal{F}$ . For simplicity, consider pure distillation, where the imitation parameter is set to $\lambda =1$.

According to VC theory \cite{vapnik1999overview}, the classification error of the classifier, $f_{s}^{b}$ can be expressed as:

\begin{equation}
R(f_{s}^{b})-R(f)\le O\left( \frac{{{\left| \mathcal{F}_{s}^{b} \right|}_{C}}}{\sqrt{n}} \right)+{{\varepsilon }_{sb}},
\label{eq:07}
\end{equation}

where the $O(\cdot )$  and ${{\varepsilon }_{sb}}$ terms are the estimation and approximation error, respectively. The former refers to the performance gap between a model on training data and its theoretical best performance. The latter refers to the difference between a model output and the true target function. It reflects whether the model representational capacity is sufficiently powerful to accurately approximate the true target function. If the hypothesis space of a model cannot capture the complexity of the target function, the approximation error will be large. Here, $R$ represents the error, ${{\left| \cdot \right|}_{C}}$ denotes a measure of the capacity of some function class, and $n$ represents the number of data points.

Let $f_{t}^{a}\in \mathcal{F}_{t}^{a}$ and $f_{t}^{b}\in \mathcal{F}_{t}^{b}$ be the teacher function trained on  $x_{i}^{a}$ and $x_{i}^{a}$ , then:

\begin{equation}
R(f_{t}^{a})-R(f)\le O\left( \frac{{{\left| \mathcal{F}_{t}^{a} \right|}_{C}}}{n} \right)+{{\varepsilon }_{ta}},
\label{eq:08}
\end{equation}

\begin{equation}
R(f_{t}^{b})-R(f)\le O\left( \frac{{{\left| \mathcal{F}_{t}^{b} \right|}_{C}}}{n} \right)+{{\varepsilon }_{tb}}.
\label{eq:09}
\end{equation}

Then, we can transfer the knowledge of the teacher separately from data `A' or `B' to the student. Let $f_{t}^{a}$  serve as the teacher function in cross-modal KD, and $f_{t}^{b}$ in Unimodal KD, then:

\textit{\textbf{Cross modal KD}}

\begin{equation}
R(f_{s}^{b})-R(f_{t}^{a})\le O\left( \frac{{{\left| \mathcal{F}_{s}^{b} \right|}_{c}}}{{{n}^{\alpha }}} \right)+{{\varepsilon }_{m}},
\label{eq:10}
\end{equation}

where  ${{\varepsilon }_{m}}$ is the approximation error of the teacher function class $\mathcal{F}_{s}^{b}$ with respect to $f_{t}^{a}\in \mathcal{F}_{t}^{a}$ , and $\frac{1}{2}\le \alpha \le 1$.

\textit{\textbf{Unimodal KD}}
\begin{equation}
R(f_{s}^{b})-R(f_{t}^{b})\le O\left( \frac{{{\left| \mathcal{F}{s}^{b} \right|}_{C}}}{{{n}^{\alpha }}} \right)+{{\varepsilon }_{l}},
\label{eq:11}
\end{equation}

where  ${{\varepsilon }_{l}}$ is the approximation error of the teacher function class $\mathcal{F}_{s}^{b}$  with respect to  $f_{t}^{b}\in \mathcal{F}_{t}^{b}$.

Then, if we employ cross-modal KD, combining Eqs. (\ref{eq:08}) and (\ref{eq:10}), we can obtain an alternative expression for the student learning the real function $f$, as follows:

\textit{\textbf{Alternative Expression Through Cross-Modal KD}}

\begin{equation}
\begin{aligned}
  & R(f_{s}^{b})-R(f)=R(f_{s}^{b})-R(f_{t}^{a})+R(f_{t}^{a})-R(f) \\ 
  & \hspace{2.1cm} \le O\left( \frac{{{\left| \mathcal{F}_{s}^{b} \right|}_{C}}}{{{n}^{\alpha }}} \right) +{{\varepsilon }_{m}}+O\left( \frac{{{\left| \mathcal{F}_{t}^{a} \right|}_{C}}}{n} \right)+{{\varepsilon }_{ta}}.
\end{aligned}
\label{eq:12}
\end{equation}

Similarly, by employing unimodal KD and combining Eqs. (\ref{eq:09}) and (\ref{eq:11}), we can obtain an alternative expression for the student's learning of the real function $f$, as follows:

\textit{\textbf{Alternative Expression Through Unimodal KD}}

\begin{equation}
\begin{aligned}
  & R(f_{s}^{b})-R(f)=R(f_{s}^{b})-R(f_{t}^{b})+R(f_{t}^{b})-R(f) \\ 
  & \hspace{2.1cm} \le O\left( \frac{{{\left| \mathcal{F}_{s}^{b} \right|}_{C}}}{{{n}^{\alpha }}} \right) +{{\varepsilon }_{l}}+O\left( \frac{{{\left| \mathcal{F}_{t}^{b} \right|}_{C}}}{n} \right)+{{\varepsilon }_{tb}}.
\end{aligned}
\label{eq:13}
\end{equation}

\textbf{Combining Eqs. (\ref{eq:12}) and (\ref{eq:13}), it is necessary to satisfy Eq. (\ref{eq:14}); otherwise, cross-modal KD would not outperform Unimodal KD.}

\begin{equation}
\underbrace{O\left( \frac{{{\left| \mathcal{F}_{t}^{a} \right|}_{C}}}{n} \right)}_{estimation}+\underbrace{{{\varepsilon }_{m}}+{{\varepsilon }_{ta}}}_{approximation}\le \underbrace{O\left( \frac{{{\left| \mathcal{F}_{t}^{b} \right|}_{C}}}{n} \right)}_{estimation}+\underbrace{{{\varepsilon }_{l}}+{{\varepsilon }_{tb}}}_{approximation}.
\label{eq:14}
\end{equation}

Observing Eq. (\ref{eq:14}), we can see that it consists of two components: estimation and approximation error. Next, we will analyze these two parts separately. 

Regarding estimation error, it is primarily determined by model capacity and data pattern strength. According to the assumptions in Sec. \ref{subsec:3.1}, when the model capacity and data strength are the same, $O(\frac{{{\left| F_{t}^{a} \right|}_{C}}}{n})$ and $O(\frac{{{\left| F_{t}^{b} \right|}_{C}}}{n})$ are equivalent. Therefore, we can eliminate $O(\frac{{{\left| F_{t}^{a} \right|}_{C}}}{n})$ and $O(\frac{{{\left| F_{t}^{b} \right|}_{C}}}{n})$ from both sides of Eq. (\ref{eq:14}) without altering its outcome. 

Regarding the approximation error term, it reflects the difference between the output of the neural network or the KD model and the target. \textbf{In a neural network model}, the model has only one objective: to minimize the difference between the network's predictions and the one-hot encoded target. Here, ${{\varepsilon }_{ta}}$ and ${{\varepsilon }_{tb}}$ represent the approximation errors when training the neural network model under the modality $a$ and the modality $b$, respectively. Under the conditions of Assumption 2, when the intensities of the data modalities are the same, the error between the prediction results trained under modality a and modality b and the ground truth are the same. Therefore, we can further eliminate ${{\varepsilon }_{ta}}$ and ${{\varepsilon }_{tb}}$ on both sides of Eq. (\ref{eq:14}). \textbf{In the KD model}, the model has two objectives: first, to minimize the discrepancy between the student output and the one-hot target; second, to reduce the difference between teacher and student outputs. ${{\varepsilon }_{l}}$ represents the unimodal KD model, while ${{\varepsilon }_{m}}$ represents the multimodal KD model. For the error generated by the first objective, regardless of whether the prediction results come from the unimodal or multimodal model, they are compared with the ground truth. When both Assumption 1 and Assumption 2 are satisfied, the approximation errors of the models is the same and can therefore be offset against each other. For the error generated by the second objective, due to the different targets of the unimodal and multimodal models (the unimodal model aims to approximate the prediction results under modality $a$, while the multimodal model aims to approximate the prediction results under modality $b$), they cannot be offset against each other. For the unimodal model, if the training time is sufficient and Assumption 1 is met, the error will approach zero because the input modalities of the teacher and student networks are the same. In contrast, due to the modality differences in the multimodal model, the error is larger and cannot be ignored.

Based on the above analysis, we simplify Eq. (\ref{eq:14}) by canceling $O(\frac{{{\left| F_{t}^{a} \right|}_{C}}}{n})$ with $O(\frac{{{\left| F_{t}^{b} \right|}_{C}}}{n})$ and ${{\varepsilon }_{ta}}$ with ${{\varepsilon }_{tb}}$ on both sides, ignoring ${{\varepsilon }_{l}}$, and retaining ${{\varepsilon }_{m}}$, ultimately reducing Eq. (\ref{eq:14}) to:

\begin{equation}
{{\varepsilon }_{m}}<0.
\label{eq:15}
\end{equation}

Since ${{\varepsilon }_{m}}$ uses the KL divergence for optimization in cross-modal KD, and according to Gibbs' inequality, the KL divergence is nonnegative, ${{\varepsilon }_{m}}$ must also satisfy the following:

\begin{equation}
{{\varepsilon }_{m}}=KL(F_{t}^{a}||F_{s}^{b})\ge 0.
\label{eq:16}
\end{equation}

By comparing Eq. (\ref{eq:15}) with Eq. (\ref{eq:16}), we arrive at a contradictory conclusion, indicating that the condition for cross-modal KD to outperform unimodal KD cannot hold. In other words, when both Assumption 1 and Assumption 2 are satisfied, the performance of cross-modal KD cannot exceed that of unimodal KD. Thus, we obtain the lower bound for the approximation error in cross-modal KD, which corresponds to unimodal KD. In this case, unimodal KD can be regarded as a special case of cross-modal KD, where there are no modality differences.

Next, we will further derive the upper bound of the approximation error ${{\varepsilon }_{m}}$ in cross-modal KD. According to \cite{zhao2022decoupled}, KL divergence can be decomposed into a part related to the target class distribution and a part related to the non-target class distribution, denoted by $KL({{p}^{ta}}||{{p}^{sb}})$ and $KL({{\hat{p}}^{ta}}||{{\hat{p}}^{sb}})$ respectively. The decomposed result is represented as:
\begin{equation}
\begin{aligned}
  & {{\varepsilon }_{m}}=KL(F_{t}^{a}||F_{s}^{b})=p_{k}^{ta}log(\frac{p_{k}^{ta}}{p_{k}^{sb}})+\sum\limits_{i=1,i\ne k}^{C}{p_{i}^{ta}}log(\frac{p_{i}^{ta}}{p_{i}^{sb}}) \\ 
 & =p_{k}^{ta}log(\frac{p_{k}^{ta}}{p_{k}^{sb}})+p_{\backslash k}^{ta}\sum\limits_{i=1,i\ne k}^{C}{\hat{p}_{i}^{ta}}(log(\frac{\hat{p}_{i}^{ta}}{\hat{p}_{i}^{sb}})+log(\frac{p_{\backslash k}^{ta}}{p_{\backslash k}^{sb}})) \\ 
 & =\underbrace{p_{k}^{ta}log(\frac{p_{k}^{ta}}{p_{k}^{sb}})+p_{\backslash k}^{ta}log(\frac{p_{\backslash k}^{ta}}{p_{\backslash k}^{sb}})}_{KL({{p}^{ta}}||{{p}^{sb}})}+p_{\backslash k}^{ta}\underbrace{\sum\limits_{i=1,i\ne k}^{C}{\hat{p}_{i}^{ta}log(\frac{\hat{p}_{i}^{ta}}{\hat{p}_{i}^{sb}})}}_{KL({{{\hat{p}}}^{ta}}||{{{\hat{p}}}^{sb}})} \\ 
 & =KL({{p}^{ta}}||{{p}^{sb}})+p_{\backslash k}^{ta}\cdot KL({{{\hat{p}}}^{ta}}||{{{\hat{p}}}^{sb}}) \\ 
\end{aligned},
\label{eq:17}
\end{equation}

where \\
\[
\begin{array}{rl}
% p_{k}^{ta} & = \sigma ({{f}_{t}}(x_{k}^{a})) = \frac{\exp ({{f}_{t}}(x_{k}^{a}))}{\sum\limits_{j=1}^{C}{\exp ({{f}_{t}}(x_{j}^{a}))}} \\
% p_{k}^{sb} & = \sigma ({{f}_{s}}(x_{k}^{b})) = \frac{\exp ({{f}_{s}}(x_{k}^{b}))}{\sum\limits_{j=1}^{C}{\exp ({{f}_{s}}(x_{j}^{b}))}} \\
p_{i}^{ta} & = \sigma ({{f}_{t}}(x_{i}^{a})) = \frac{\exp ({{f}_{t}}(x_{i}^{a}))}{\sum\limits_{j=1}^{C}{\exp ({{f}_{t}}(x_{j}^{a}))}}, \\
p_{i}^{sb} & = \sigma ({{f}_{s}}(x_{i}^{b})) = \frac{\exp ({{f}_{s}}(x_{i}^{b}))}{\sum\limits_{j=1}^{C}{\exp ({{f}_{s}}(x_{j}^{b}))}}, \\
p_{\backslash k}^{ta} & = \sigma ({{f}_{\backslash k}}(x_{i}^{a})) = \frac{\sum\limits_{i=1, i\ne k}^{C}{\exp ({{f}_{t}}(x_{i}^{a}))}}{\sum\limits_{j=1}^{C}{\exp ({{f}_{t}}(x_{j}^{a}))}}. \\
% \hat{p}_{i}^{ta} & = \sigma ({{f}_{t}}(\hat{x}_{i}^{a})) = \frac{\exp ({{f}_{t}}(\hat{x}_{i}^{a}))}{\sum\limits_{j=i, j\ne k}^{C}{\exp ({{f}_{t}}(x_{j}^{a}))}} \\
% \hat{p}_{i}^{sb} & = \sigma ({{f}_{s}}(\hat{x}_{i}^{b})) = \frac{\exp ({{f}_{s}}(\hat{x}_{i}^{b}))}{\sum\limits_{j=i, j\ne k}^{C}{\exp ({{f}_{s}}(x_{j}^{b}))}}
\end{array}
\]

Based on Eq. (\ref{eq:17}), the upper bound of the approximation error in cross-modal KD can be obtained as follows:

\begin{equation}
\begin{aligned}
  & {{\varepsilon }_{m}}=KL({{p}^{ta}}||{{p}^{sb}})+p_{\backslash k}^{ta}\cdot KL({{{\hat{p}}}^{ta}}||{{{\hat{p}}}^{sb}}) \\ 
 & \le \underbrace{KL({{p}^{ta}}||{{p}^{sb}})}_{\text{TCKL}}+\underbrace{KL({{{\hat{p}}}^{ta}}||{{{\hat{p}}}^{sb}})}_{\text{NCKL}}. \\ 
\end{aligned}
\label{eq:18}
\end{equation}

According to Eq. (\ref{eq:18}), the upper bound of the approximation error in cross-modal KD is composed of two parts: one part is determined by the distribution error of the target classes, referred to as TCKL; the other part is determined by the distribution error of the non-target classes, referred to as NCKL. Compared to TCKL, NCKL involves the probability distributions of more classes, thereby increasing its complexity and uncertainty. It can be proven that when the number of non-target classes exceeds two, NCKL becomes significantly larger than TCKL. To demonstrate this, consider the expressions for TCKL and NCKL:

\begin{equation} \text{TCKL}=p_{k}^{ta}\log\left(\frac{p_{k}^{ta}}{p_{k}^{sb}}\right) \label{eq:19}, \end{equation}

\begin{equation} \text{NCKL}=\sum\limits_{i=1,i\ne k}^{C}\hat{p}_{i}^{ta}\log\left(\frac{\hat{p}_{i}^{ta}}{\hat{p}_{i}^{sb}}\right). \label{eq:20} \end{equation}

The ratio of NCKL to TCKL can be derived from Eqs. (\ref{eq:19}) and (\ref{eq:20}), as shown in

\begin{equation} \frac{\text{NCKL}}{\text{TCKL}}=\frac{\sum\limits_{i=1,i\ne k}^{C}{\hat{p}_{i}^{ta}log(\frac{\hat{p}_{i}^{ta}}{\hat{p}_{i}^{sb}})}}{p_{k}^{ta}log(\frac{p_{k}^{ta}}{p_{k}^{sb}})}. \label{eq:21} \end{equation}

To estimate \(\mathrm{NCKL/TCKL}\), we perform a Taylor series expansion and approximate by neglecting higher-order infinitesimal terms. Let $\hat{p}_{i}^{sb}=\hat{p}_{i}^{ta}+{\varepsilon}_{i}$ and $p_{k}^{sb}=p_{k}^{ta}+{\varepsilon }_{t}$, where ${\varepsilon }_{i}$ and ${\varepsilon }_{t}$ are small quantities. After the Taylor series expansion, the result of $\log \left( \frac{\hat{p}{i}^{ta}}{\hat{p}_{i}^{sb}} \right)$ is as follows:

\begin{equation} \log \left( \frac{\hat{p}_{i}^{ta}}{\hat{p}_{i}^{sb}} \right)\approx \log \left( 1-\frac{{\varepsilon}_{i}}{\hat{p}_{i}^{ta}} \right)\approx -\frac{{\varepsilon }_{i}}{\hat{p}{i}^{ta}}-\frac{1}{2}\left( \frac{{\varepsilon }_{i}}{\hat{p}_{i}^{ta}} \right)^{2}. \label{eq:22} \end{equation}

Therefore, $\text{NCKL}$ and $\text{TCKL}$ can be approximated by:

\begin{align}
\text{NCKL} &\approx \sum\limits_{i\ne k}{\hat{p}_{i}^{ta}}\left( -\frac{{{\varepsilon }_{i}}}{\hat{p}_{i}^{ta}}-\frac{1}{2}{{\left( \frac{{{\varepsilon }_{i}}}{\hat{p}_{i}^{ta}} \right)}^{2}} \right) \notag \\
&= -\sum\limits_{i\ne k}{{{\varepsilon }_{i}}}-\frac{1}{2}\sum\limits_{i\ne t}{\frac{\varepsilon _{i}^{2}}{\hat{p}_{i}^{ta}}},
\label{eq:23}
\end{align}

\begin{align}
\text{TCKL}\approx p_{k}^{ta}\log \left( \frac{p_{k}^{ta}}{p_{k}^{ta}+{{\varepsilon }_{t}}} \right)\approx -p_{k}^{ta}\frac{{{\varepsilon }_{t}}}{p_{k}^{ta}}=-{{\varepsilon }_{t}}.
\label{eq:24}
\end{align}

From Eqs. (\ref{eq:23}) and (\ref{eq:24}), \(\mathrm{NCKL/TCKL}\) can be approximated by:

\begin{equation} \frac{\text{NCKL}}{\text{TCKL}}\approx \frac{-\sum\limits_{i\ne k}{{{\varepsilon }_{i}}}-\frac{1}{2}\sum\limits_{i\ne k}{\frac{\varepsilon _{i}^{2}}{\hat{p}_{k}^{ta}}}}{-{{\varepsilon }_{t}}}\ge \frac{C-1}{1}\cdot \frac{{{\varepsilon }_{i}}}{{{\varepsilon }_{t}}}. \label{eq:25} \end{equation}

Based on Assumptions 1 and 2, the prediction errors of the teacher and student networks on the target class are comparable, so ${\varepsilon }_{t}$ is much smaller than ${\varepsilon }_{i}$. When $C-1$ is sufficiently large, \(\mathrm{NCKL/TCKL}\) will be much greater than 1. In summary, we have proven that in cross-modal KD, when the number of classes $C$ is greater than 2, $\text{NCKL}$ is significantly larger than $\text{TCKL}$. Therefore, the distribution differences among the non-target classes are the key factors that influence the effectiveness of cross-modal KD. When the distribution differences among non-target classes decrease, the upper bound of the approximation error will correspondingly reduce, thereby enhancing the effectiveness of cross-modal KD.

\section{Experiments}

\begin{figure*}[!t]
\centering
\includegraphics[width=0.95\linewidth]{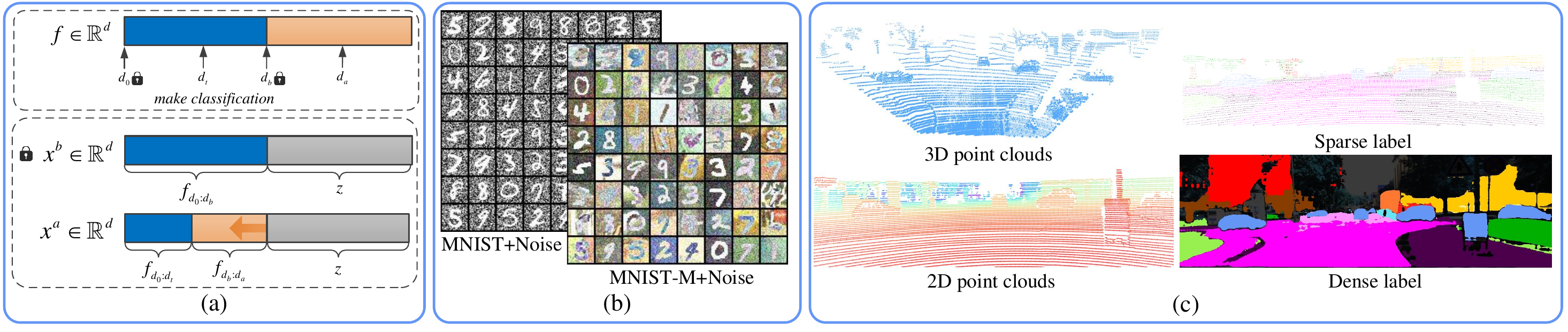}
\caption{Data Processing. (a) Scikit-learn Data: With the student modality ${{x}^{b}}$ fixed, the teacher modality ${{x}^{a}}$ is adjusted by changing the parameter $\gamma $, which ranges from 0 to 1. A larger $\gamma $ indicates a greater domain difference between the modalities. (b) MNIST/MNIST-M: Random noise is added to the MNIST and MNIST-M datasets to test the robustness of the model. (c) SemanticKITTI: 3D point cloud data is projected into the camera coordinate system to produce 2D point clouds, and dense image segmentation labels are obtained as described in \cite{chen2024foundation}.}
\label{fig:4}
\end{figure*}

In this section, we validate the hypothesis through experiments conducted on five multimodal datasets. First, we introduce the datasets and network models used in the experiments. Next, we elaborate on the experimental design and methodology. Finally, we assess the validity of the hypothesis in the context of traditional KD \cite{hinton2015distilling}. Additionally, we demonstrate the broad applicability and practical value of NTDH.

\subsection{Datasets}

The five multimodal datasets encompass three types: a simulated dataset (Scikit-learn), a synthetic dataset (MNIST/MNIST-M), and three real-world datasets (RAVDESS, SemanticKITTI, and NYU Depth V2). To ensure that the experimental datasets adhere to the assumptions outlined in Section \ref{subsec:3.1} and to mitigate the effects of data modality strength and model capacity on the results, we perform preprocessing. The adjustments to model capacity and modality strength are summarized in Table \ref{tab:1}. 

\begin{table}[]
\centering
\caption{Model Capacity and Modality Strength. Model capacity refers to the number of parameters. Modality strength refers to the test performance. Specifically, the Scikit-learn, MNIST/MNIST-M, and RAVDESS datasets report classification accuracy, while the semanticKITTI and NYU-Depth V2 datasets report the mean Intersection over Union (mIoU) for segmentation tasks.}
\label{tab:1}
\begin{tabular}{lcc}
\toprule
Dataset       & Model Capacity    & Modality   Strength \\ \midrule

RAVDESS       & \makecell{Image: 13.07M \\ Audio: 13.24M} & \makecell{Image: 71.4\% \\ Audio: 74.6\%} \\
% \rowcolor{gray!20} % 设置 SemanticKITTI 第一行的背景色

\rowcolor{gray!20} % 设置 MNIST/MNIST-M 行的背景色
Scikit-learn Data      & 0.0014M          & 39.6\%               \\

\multirow{2}{*}{semanticKITTI} & \multirow{2}{*}{21.42M} & Lidar: 60.5\% \\
% \rowcolor{gray!20} % 设置 SemanticKITTI 第二行的背景色
                              &                    & Image: 58.4\% \\
\rowcolor{gray!20} % 设置 MNIST/MNIST-M 行的背景色                              
MNIST/MNIST-M & 0.53M            & 69.1\%              \\
\multirow{2}{*}{NYU-Depth V2}  & \multirow{2}{*}{0.17M} & Depth: 24.7\% \\

                              &                    & Image: 24.4\% \\
\bottomrule
\end{tabular}
\end{table}

% Specifically, to satisfy Assumption 1, the dimensions of the multimodal data, except for the channel dimension, were unified so that the same network could be used for feature extraction across different modalities, ensuring consistency in model capacity. Additionally, to meet Assumption 2, noise is added to the MNIST/MNIST-M, and dense labels were added to the RGB data in the SemanticKITTI, ensuring that the accuracy of different modalities under the same model is comparable, thus balancing the strength of the data modalities. The results of the modified model capacity and modality strength are shown in Table \ref{tab:1}. The details of the datasets are as follows:

\subsubsection{Scikit-learn}

The Scikit-learn dataset is generated using the Scikit-learn toolkit, which allows us to precisely control the differences and strengths between modalities, ensuring that the data fully meets \textit{Assumptions} 1 and 2. Specifically, we use the \textit{make\_classification} function from the toolkit to generate multi-class data, where the labels are determined by the feature $f$, and the length of the feature is $d$. Next, let ${{x}^{a}}\in {{\mathbb{R}}^{d}}$, ${{x}^{b}}\in {{\mathbb{R}}^{d}}$, and $y$ constitute the multimodal data $({{x}^{a}},{{x}^{b}},y)$. Each modality is formed by a portion of the decisive features of $x$ combined with noise $z\in {{\mathbb{R}}^{{{d}_{z}}}}$. Specifically, ${{x}^{b}}=\left[ {{x}_{{{d}_{0}}:{{d}_{b}}}},z \right]\in {{\mathbb{R}}^{d}}$, ${{x}^{a}}=\left[ {{x}_{{{d}_{0}}:{{d}_{t}}}},{{x}_{{{d}_{b}}:{{d}_{a}}}},z \right]\in {{\mathbb{R}}^{d}}$, and ${{d}_{a}}+{{d}_{t}}=2\cdot {{d}_{b}}$. We define a ratio $\gamma ={{{d}_{a}}-{{d}_{b}}}/{{{d}_{b}}}\;\in [0,1]$, which characterizes the ratio of domain difference between modalities. As $\gamma $ increases, the domain difference between data modalities also increases. In the experiment, ${{x}^{b}}\in {{\mathbb{R}}^{{{d}}}}$ is fixed, and $\gamma $ is set to $\gamma =[0,0.25,0.5,0.75,1]$, constructing data with varying modality domain differences, as shown in Fig. \ref{fig:4}(a).

\subsubsection{MNIST/MNIST-M} 
MNIST \cite{lecun1998gradient} is a widely used dataset for handwritten digit recognition, containing grayscale images of digits from 0 to 9, each paired with a corresponding label. Each image has a resolution of 28x28 pixels, with the dataset comprising 60,000 training images and 10,000 test images. MNIST-M \cite{ganin2015unsupervised} is derived from the MNIST digits by randomly mixing colored patches from the BSDS500 \cite{martin2001database}, creating a different modality of the same handwritten digits. Since the accuracy performance of MNIST and MNIST-M varies under the same model, to satisfy Assumption 2, which requires that the data strengths of both modalities be the same, we add noise of varying intensities into MNIST and MNIST-M, as shown in Fig. \ref{fig:4}(b).

\subsubsection{RAVDESS}
The RAVDESS \cite{livingstone2018ryerson} is a dataset used for multimodal emotion recognition, containing data in both audio and video modalities. The dataset consists of speech and song recordings by 24 actors (12 male and 12 female), covering emotional categories such as neutral, happy, angry, fearful, disgusted, surprised, and sad. RAVDESS includes a total of 1,440 files and is widely used in research on emotion recognition, multi-modal learning, and human-computer interaction systems. In our study, we use the video modality as the input for the student network, while the audio modality serves as the input for the teacher network.

\subsubsection{SemanticKITTI}

SemanticKITTI \cite{behley2019semantickitti} is a large-scale dataset based on the KITTI odometry benchmark, providing 43,000 scans with point-wise semantic annotations, of which 21,000 scans (sequences 00-10) are available for training and validation. The dataset includes 19 semantic categories and is used for the evaluation of point cloud semantic segmentation benchmarks. To enable different modalities to be processed by the same network, we follow \cite{zhuang2021perception} and project the original 3D point cloud data into the camera coordinate system, resulting in 2D point cloud features, as shown on the left side of Fig. \ref{fig:4}(c).

Additionally, in point cloud semantic segmentation, the image modality lacks dense semantic labels, leading to significantly higher segmentation accuracy for the point cloud modality compared to the image modality. To equalize the data strength, we obtain dense semantic labels for the image modality following the method described in \cite{chen2024foundation}, as shown on the right side of Fig. \ref{fig:4}(c).

\subsubsection{NYU Depth V2} 
The NYU Depth V2 dataset, introduced by Silberman et al. in 2012, is collected using Microsoft Kinect's RGB and depth cameras from commercial and residential buildings across three cities in the United States. The dataset consists of 1,449 densely annotated pairs of RGB and depth images, covering 464 distinct indoor scenes across 26 scene categories. In this study, the RGB data serve as the input for the teacher network, while the depth images are used as input for the student network, and experiments are conducted in the context of the semantic scene completion task.

\subsection{Network Model}
We fine-tune the existing network models to serve as the teacher and student models for our experiments. Specifically, for the Scikit-learn, both the teacher and student networks use a 2-layer fully connected network; for the MNIST/MNIST-M, both employ a 3-layer fully connected network; for the RAVDESS, due to the differing shapes of audio and image data, both the teacher and student networks utilize a 2D convolutional network with the same number of layers, and the kernel sizes are adjusted to accommodate the different data modalities. In the SemanticKITTI, the overall network architecture proposed by \cite{zhuang2021perception} is adopted, with the intermediate fusion layer removed; similarly, in the NYU Depth V2, the network architecture proposed by \cite{li2020anisotropic} is used, and the fusion layer is also removed.

\subsection{Experiment Plan}
\begin{figure}[!htbp]
\centering
\includegraphics[width=0.95\linewidth]{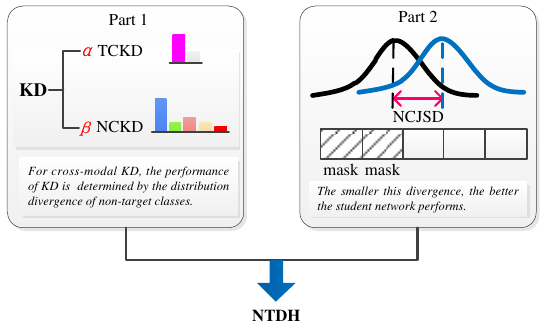}
\caption{The experimental plan for NTDH. The first part involves adjusting weight coefficients in the loss function to assess the impact of distribution differences among non-target classes on KD. The second part applies a masking method to selectively remove features or samples with significant non-target distribution differences, evaluating the subsequent effect on distillation performance.}
\label{fig:5}
\end{figure}

To verify the validity of our hypothesis, we design a detailed experimental plan. We employ a divide-and-conquer approach by splitting the hypothesis into two parts and validating each separately. For the first part, we use a weight adjustment method to verify that the distribution differences among non-target classes are the primary factors affecting the effectiveness of KD. This is done by adjusting the weight coefficients in the loss function that account for the distribution differences between target and non-target classes. For the second part, we apply a masking method to remove features or samples that have a significant impact on the distribution differences among non-target classes, observing whether this can improve the performance of cross-modal KD. The experimental plan is illustrated in Fig. \ref{fig:5}, with the detailed steps as follows:

\subsubsection{The First Part} We observe whether the distribution differences among non-target classes are the primary factors affecting KD by altering the weights in the loss function that optimize the distribution differences between target and non-target classes. We refer to this method as the weight adjustment method. Specifically, we first decompose the KL divergence-based KD loss function into TCKL and NCKL according to Eqs. (\ref{eq:19}) and (\ref{eq:20}). TCKL aims to reduce the distribution gap between the teacher and student for target classes, while NCKL focuses on reducing the distribution gap for non-target classes. Next, we assign weight coefficients $\alpha$ and $\beta$ to TCKL and NCKL, respectively. Finally, by adjusting these weight coefficients, we either increase or decrease the influence of the distribution divergence of non-target classes.

\subsubsection{The Second Part} In the first part of the study, we confirm that the distribution differences among non-target classes are a key factor affecting the effectiveness of cross-modal KD. To further explore the relationship between non-target distribution differences and KD performance, we propose a \textit{mask-based method}. This method ranks the features or samples based on the non-target distribution differences. After ranking, we only use the top-ranked features or samples during the distillation process to reduce the non-target distribution differences. If this method improves the performance of KD, it validates our hypothesis. For smaller datasets, such as the Scikit-learn dataset, MNIST/MNIST-M, and RAVDESS, we apply the \textit{feature mask}, while for larger datasets, such as SemanticKITTI and NYU Depth V2, we use the \textit{sample mask}. The specific algorithm is as follows:

\begin{algorithm}[H]
\caption{Feature Mask}
\begin{algorithmic}[1]
\State \textbf{Input:} ${{x}^{a}}\in {{\mathbb{R}}^{n\times {{d}_{1}}}}$, ${{x}^{b}}\in {{\mathbb{R}}^{n\times {{d}_{2}}}}$, and $y\in {{\mathbb{R}}^{n}}$
\State \textbf{Output:} Salience vector $p \in \mathbb{R}^{d_1}$ for features of modality $a$
\Procedure{Feature Mask}{}
\State Jointly train two unimodal networks ${{f}_{a}}$ and ${{f}_{b}}$ using the loss:
\begin{equation}
\begin{aligned}
\min L = & \ \text{Dist}({{f}_{a}}({{x}^{a}}), {{f}_{b}}({{x}^{b}})) \\
         & + \text{CE}(y, {{f}_{a}}({{x}^{a}})) + \text{CE}(y, {{f}_{b}}({{x}^{b}}))
\end{aligned}
\label{eq:26}
\end{equation}
\Comment{$\text{Dist}(\cdot ,\cdot )$ denotes a distance loss (e.g., Kullback-Leibler Divergence), and $\text{CE}$ denotes cross entropy.}
\State Calculate the salience for the $i$-th feature dimension.
\For{$i=1$ to ${{d}_{1}}$}
    \State ${{p}_{i}}=0$
    \For{$m=1$ to $M$} \Comment{Repeat permutation $M$ times for better stability}
        \State Mask the $i$-\textit{th} column of ${{x}^{a}}$
        \State ${{p}_{i}}={{p}_{i}}+\frac{1}{M}\cdot \text{NCJSD}$ \Comment{NCJSD can be obtained using Eq. (\ref{eq:06}).}
    \EndFor
\EndFor
\State Perform normalization: $p=\frac{p}{\max {{p}_{i}}}$ 
\State \textbf{return} $p$
\EndProcedure
\end{algorithmic}
\end{algorithm}

\textbf{Feature Mask:} The main steps of the feature mask method are shown in Algorithm 1. The inputs are ${{x}^{a}}\in {{\mathbb{R}}^{n\times {{d}_{1}}}}$, ${{x}^{b}}\in {{\mathbb{R}}^{n\times {{d}_{2}}}}$, and $y\in {{\mathbb{R}}^{n}}$, representing $n$ pairs of features from modalities $a$ and $b$, as well as the corresponding labels for these $n$ targets. The output is a saliency vector $p\in {{\mathbb{R}}^{{{d}_{1}}}}$, which represents the difference in non-target class distribution between the teacher and student networks, where the $i$-\textit{th} entry ${{p}_{i}}\in [0,1]$ reflects the saliency of the corresponding feature dimension. A larger saliency value indicates a greater difference in non-target class distribution for that feature channel. 

Algorithm 1 is designed based on a backtracking approach starting from the output layer. Specifically, in Step 4, we jointly train two unimodal networks, ${{f}_{a}}$ and ${{f}_{b}}$, which take unimodal data ${{x}_{a}}$ and ${{x}_{b}}$ as inputs, respectively. The first term in Eq. (\ref{eq:26}) aims to align the feature spaces learned by the two networks, while the remaining terms ensure that the learned features accurately predict the labels. In Step 5, we utilize the idea of feature importance ranking \cite{wojtas2020feature} to trace the saliency of input features  with respect to non-target class distribution differences. For the $i$-\textit{th} dimension in ${{x}_{a}}$, we randomly permute the values along that dimension and obtain the permuted result ${{\tilde{x}}_{a}}$ in Step 8. Next, in Step 10, we quantify the difference in non-target class distribution for each input feature channel by calculating the NCJSD, where a larger distance indicates a more significant difference between the teacher and student networks. Therefore, we can quantify the magnitude of the non-target class distribution difference in each input feature channel using the saliency vector $p$.

We repeat the permutation process multiple times and average the distance values to improve stability. Finally, in Step 13, $p$ is normalized to $[0,1]$. Once the feature channels are ranked, we can reduce the non-target class distribution differences during distillation by applying a mask to the top $\gamma\%$ of the feature channels.

\begin{algorithm}[H]
\caption{Sample Mask}
\begin{algorithmic}[1]
\State \textbf{Input:} $\sigma ({{f}_{t}}(x_{i}^{a}))$, $\sigma ({{f}_{s}}(x_{i}^{b}))$ \Comment{Predicted probability distributions of teacher and student networks}
\State \textbf{Output:} Saliency vector $P \in \mathbb{R}^{n}$ \Comment{Indicates difference in non-target class distribution}
\Procedure{Sample Mask}{}
\State Calculate non-target class probabilities $\hat{p}_{i}^{ta}$ and $\hat{p}_{i}^{sb}$ for teacher and student networks, excluding target class $k$.
\For{$i=1$ to $n$}
    \State Calculate the NCJSD distance for each sample according to Eq. (\ref{eq:06}). 
    
\EndFor
\State Sort the samples based on their NCJSD values in descending order.
\For{$i=1$ to $n$}
    \State Calculate the saliency $p_i$ for the $i$-\textit{th} sample based on the sorted NCJSD distances.
\EndFor
\State Normalize the saliency vector: $p = \frac{p}{\max p_i}$ \Comment{Normalize to $[0,1]^{n}$}
\State Apply a mask to the top $\gamma\%$ of samples based on the saliency vector $p$.
\State \textbf{return} $p$
\EndProcedure
\end{algorithmic}
\end{algorithm}

\textbf{Sample Mask:} The main steps of the feature masking method are illustrated in Algorithm 2. The inputs to Algorithm 2 are $\sigma ({{f}_{t}}(x_{i}^{a}))$ and $\sigma ({{f}_{s}}(x_{i}^{b}))$, representing the predicted probability distributions of the teacher and student networks, respectively. The output is a saliency vector $P\in {{\mathbb{R}}^{n}}$, which indicates the difference in non-target class distribution between the teacher and student networks, where the $i$-\textit{th} entry ${{p}_{i}}\in [0,1]$ reflects the saliency of the corresponding sample. A larger saliency value indicates that the sample exhibits a greater difference in non-target class distribution. Here, $n$ represents the number of samples, such as the number of points in each batch for the SemanticKITTI or the number of pixels in each batch for the NYU Depth V2.

In Step 6, we calculate the NCJSD for all samples in each batch. In Step 8, we sort these distances, where larger distances indicate more significant differences in non-target class distribution between the teacher and student networks. Therefore, we can quantify the magnitude of the non-target class distribution difference for each sample using the saliency vector $p$. Finally, in Step 12, $p$ is normalized to ${[0,1]}$. Once the sorting of all samples in the batch is completed, we can reduce the distribution differences between the non-target classes during distillation by applying a mask to the top $\gamma\%$ of the samples.

\subsection{The Results of the First Part}
\subsubsection{Scikit-learn}

\begin{figure}[!htbp]
\centering
\includegraphics[width=0.98\linewidth]{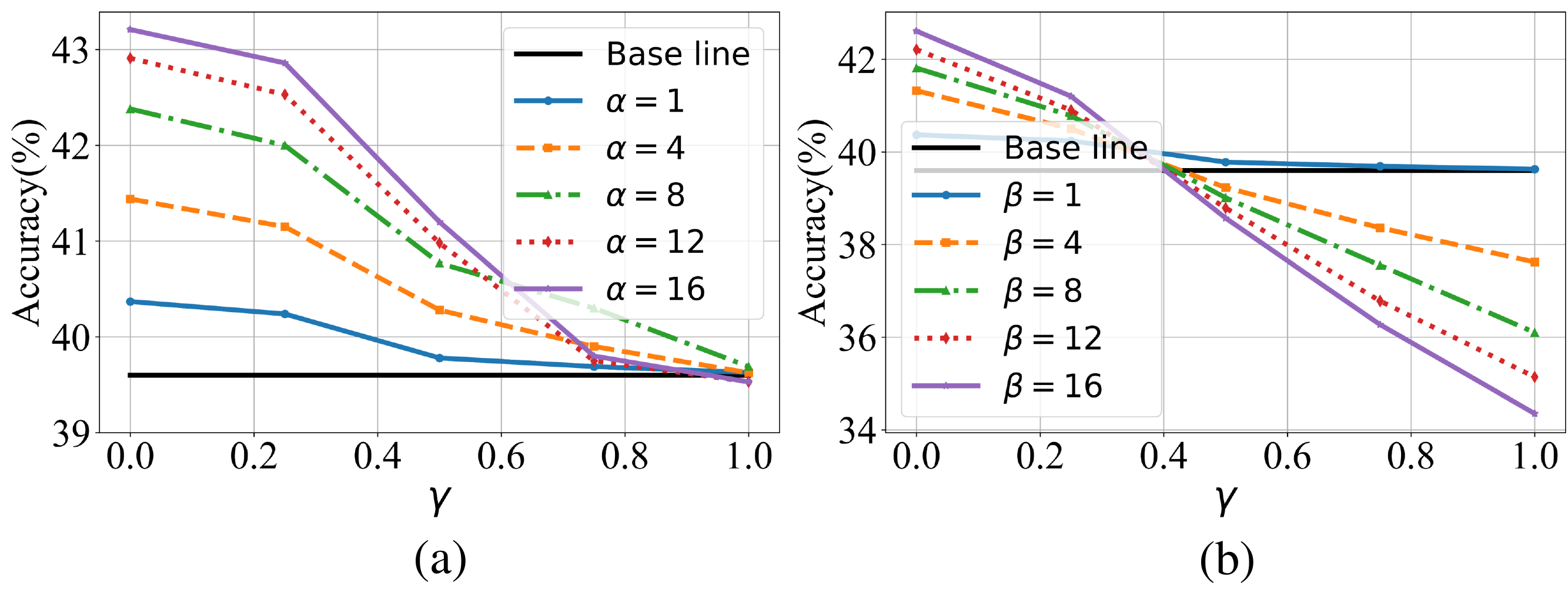}
\caption{Results of the weight adjustment method on the Scikit-learn dataset: (a) With $\beta $ fixed, as $\alpha $ increases, the weight of the target class gradually increases. As the modality difference $\gamma$ increases, the weight difference for the target class becomes minimal. (b) With $\alpha $ fixed, as $\beta $ increases, the weight for the sub-target class gradually increases. As the modality difference $\gamma$ increases, the influence of the sub-target class also gradually increases. Notably, when $\gamma >0.4$, increasing the weight of non-target classes leads to negative KD.}
\label{fig:6}
\end{figure}

The results of the Scikit-learn dataset are shown in Fig. \ref{fig:6}. The horizontal axis represents the modality difference $\gamma$, where $\gamma = 0$ indicates no difference and $\gamma = 1$ indicates the maximum difference. The experiment sets six levels of modality divergence. By adjusting the weights $\alpha$ and $\beta$, we study the impact of distribution divergence between target and non-target classes on the effectiveness of KD.

In Fig. \ref{fig:6}(a), we fix $\alpha$ and increase $\beta $ to give more weight to the target class under the six different modality differences. Correspondingly, in Fig. \ref{fig:6}(b), we fix the $\beta$ and increase the $\alpha$ to give more weight to the distribution divergence of non-target classes. By comparing the experimental results, it can be observed that when $\gamma =1$, indicating a large modality difference, increasing $\alpha$ significantly improves the effectiveness of KD, whereas increasing $\beta$ significantly reduces the effectiveness. This indicates that, in cases of large modality differences, the distribution divergence of non-target classes play a crucial role in the effectiveness of KD.

% Additionally, when $\beta$ is fixed and $\alpha$ increases, the increase in $\alpha$ initially enhances the effect of cross-modal knowledge distillation (KD) as $\gamma$ increases, but the effect gradually diminishes afterwards. In contrast, when $\alpha$ is fixed and $\beta$ increases, the increase in $\beta$ initially enhances cross-modal KD, but the effect significantly decreases as $\gamma$ continues to increase. Notably, when $\gamma >0.4$, the performance is even worse than that of the method without KD.

\subsubsection{Other dataset}

\begin{figure}[!htbp]
\centering
\includegraphics[width=0.98\linewidth]{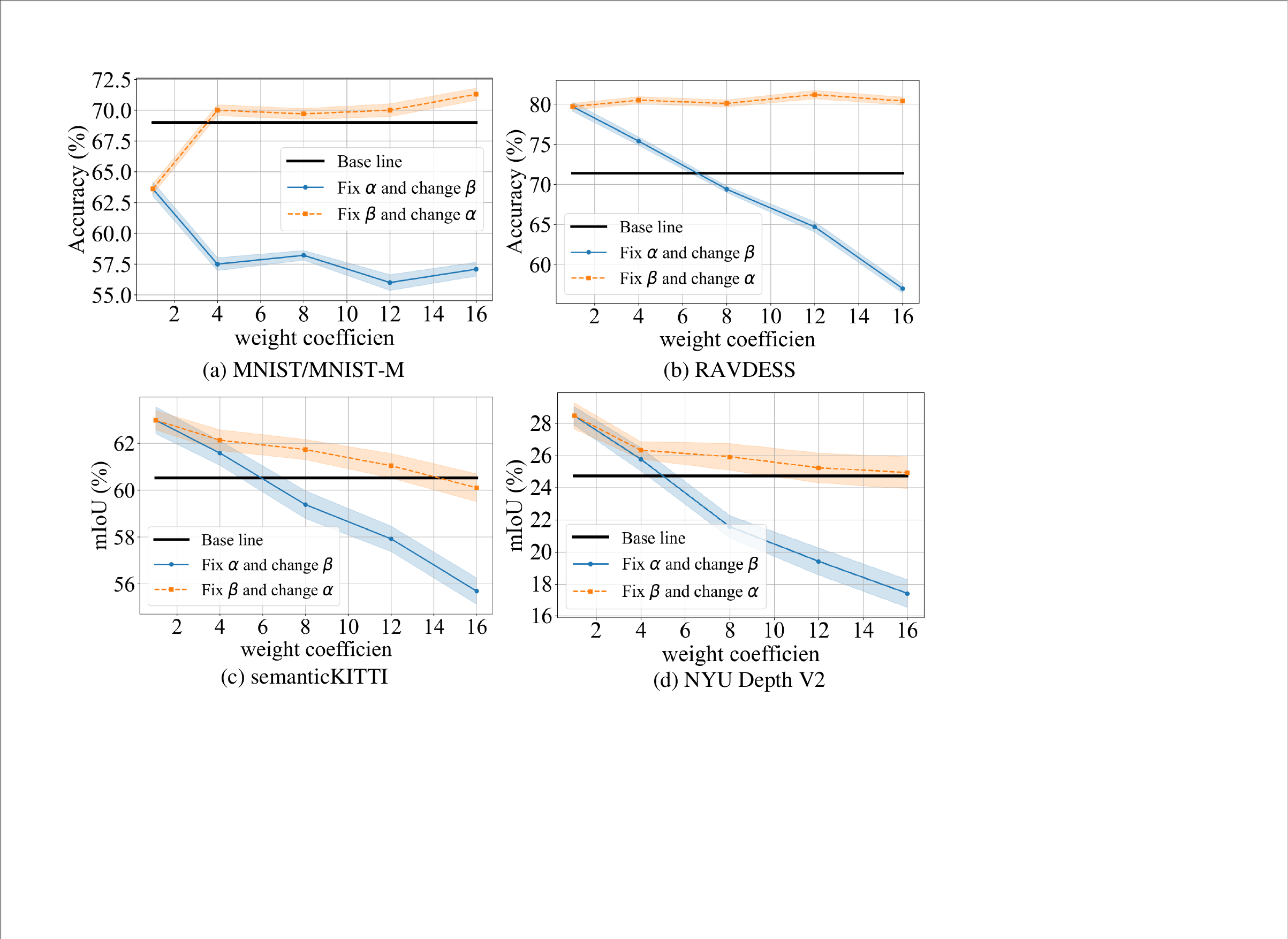}
\caption{Results of the weight adjustment method on other datasets: (a) On the MNIST/MNIST-M dataset, increasing the weight for non-target class differences significantly reduces accuracy, while adjusting the weight for target class differences has a minimal effect on accuracy. (b) to (d) show similar trends across the RAVDESS, SemanticKITTI, and NYU Depth V2 datasets, confirming that the distribution difference of non-target classes is a key factor influencing cross-modal knowledge extraction.}
\label{fig:7}
\end{figure}

On the MNIST/MNIST-M datasets, we employ the \textit{feature mask}, fixing the $\beta$ parameter while gradually increasing the $\alpha$ to amplify the weight of the distribution divergence of non-target classes. We then fix the $\alpha$ and increase the $\beta$ to enhance the weight of target class distribution differences. The results show that when the weight of non-target class distribution divergence increases, accuracy significantly decreases, while increasing the weight of target class distribution divergence results in almost no noticeable change. This indicates that the distribution divergence of non-target classes is a crucial factor influencing the effectiveness of cross-modal KD. To further validate this phenomenon, similar  experiments are conducted on the RAVDESS, SemanticKITTI, and NYU Depth V2 datasets, and consistent results are obtained, as shown in Figs. \ref{fig:7}(b)-(d), respectively. These consistent results across different datasets further confirm our hypothesis.

\subsection{The Results of the Second Part}
\subsubsection{Scikit-learn}

\begin{figure}[!htbp]
\centering
\includegraphics[width=0.95\linewidth]{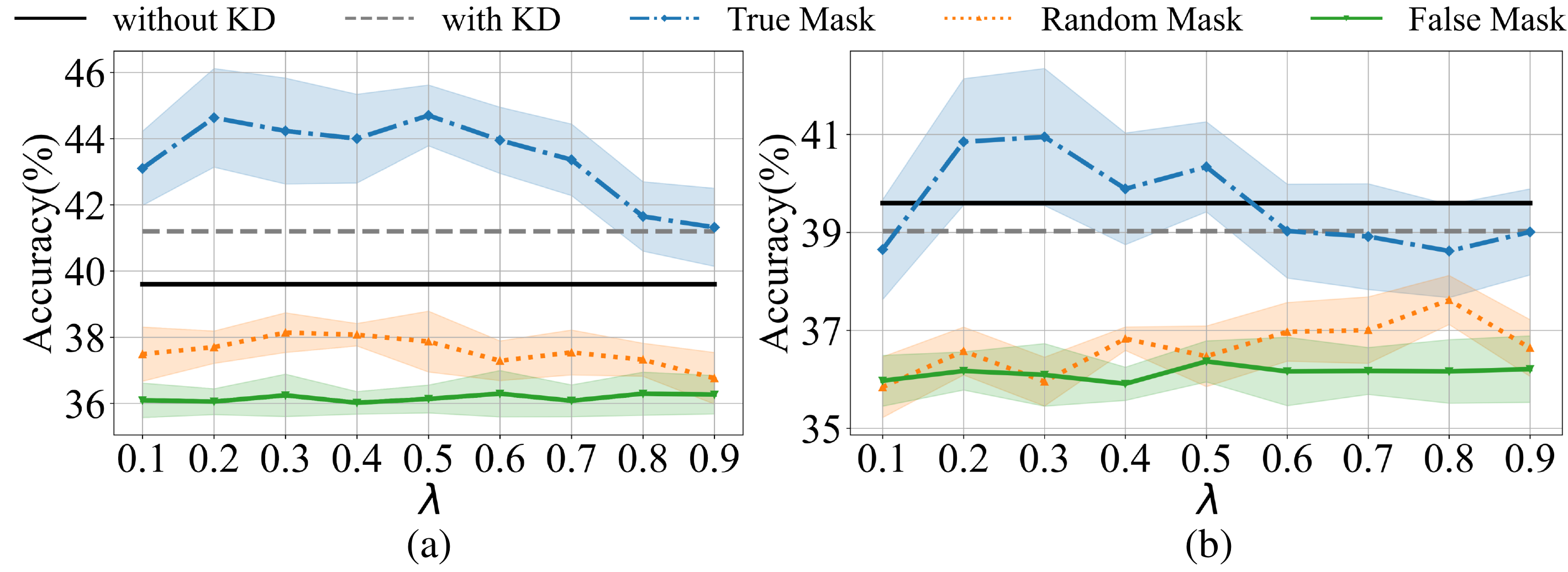}
\caption{Results of the feature masking method on the Scikit-learn dataset: (a) Results when $\gamma =0.5$. (b) Results when $\gamma =0$. The True mask reduces the
distribution divergence of non-target classes by deactivating features with the largest NCJSD, while the False mask increases these differences by deactivating features with the smallest NCJSD. The random mask deactivates features randomly. The \textit{True Mask} initially improves the performance of the student network by discarding features with larger differences, but performance decreases later as available features diminish. The \textit{False Mask} shows the worst performance, and the \textit{Random Mask} performs between the two.}
\label{fig:8}
\end{figure}

By applying Algorithm 1 to deactivate the top $\gamma$ of features with the highest NCJSD, we reduce the non-target distribution divergence between the teacher and student networks, which is referred to as the \textit{True Mask}. For control purposes, we establish two additional groups: one that deactivates the top $\gamma$ of features with the lowest NCJSD, thereby increasing the non-target distribution divergence between the teacher and student networks, is known as the \textit{False Mask}; the other randomly deactivates $\gamma$ of features, referred to as the \textit{Random Mask}.

Fig. \ref{fig:8} shows the results for $\gamma = 0.5$ and $\gamma = 0$ on the Scikit-learn dataset. The 0-\textit{th} layer features of the teacher network (i.e., the input data) are selected as the masking object, with nine different masking ratios ranging from 10\% to 90\%. For the \textit{True Mask}, a higher removal ratio indicates that more features with greater non-target distribution divergence are removed from the teacher network. Conversely, for the \textit{False Mask}, the situation is reversed. The results show that for the \textit{True Mask}, the performance of the student network initially improves with increasing $\gamma\%$. This improvement suggests that features with significant non-target distribution divergence between the teacher and student networks are being progressively discarded, thereby enhancing the performance of the student network. However, as the deactivation process starts to affect features with smaller non-target distribution divergence, performance begins to decline. In contrast, the \textit{False Mask} performs the worst, while the \textit{Random Mask} results fall between the two. This observation is consistent with our expectations.

\subsubsection{MNIST/MNIST-M and RAVDESS}

\begin{figure}[!htbp]
\centering
\includegraphics[width=0.95\linewidth]{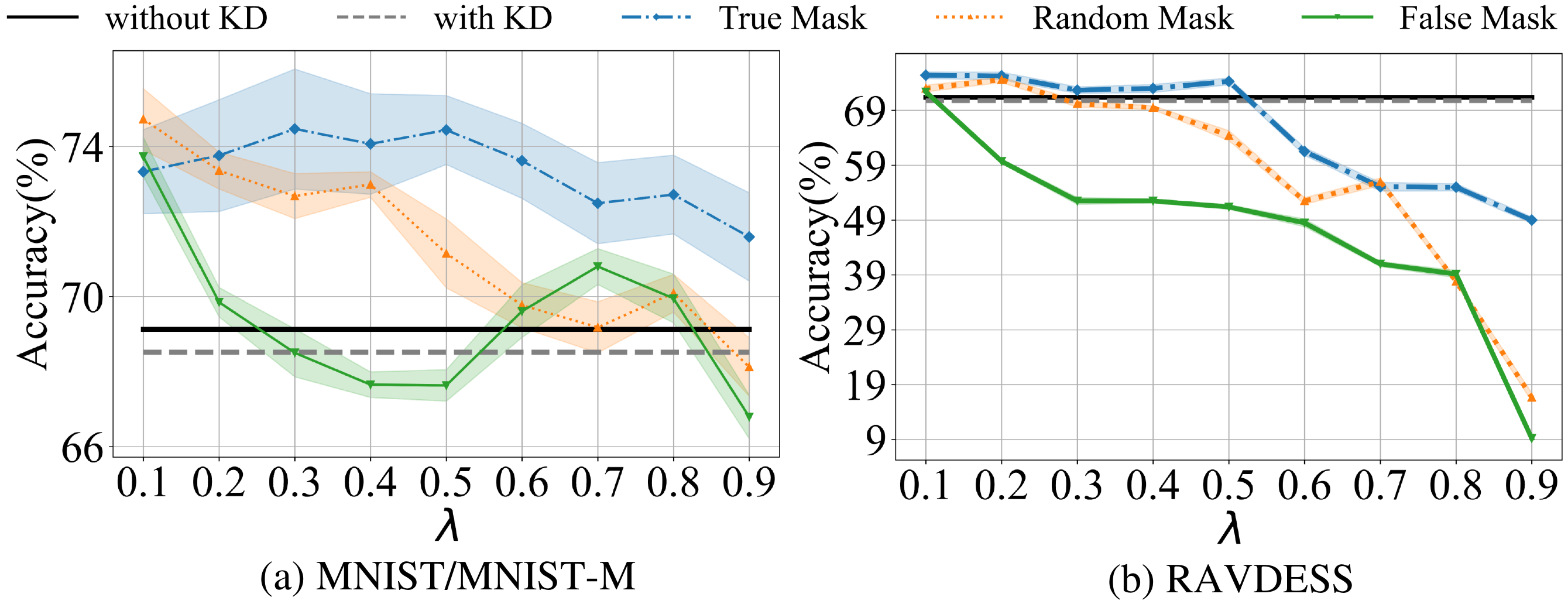}
\caption{Results of the feature masking method on the MNIST/MNIST-M and RAVDESS datasets: (a) Results on the MNIST/MNIST-M dataset; (b) Results on the RAVDESS dataset. The \textit{True Mask} initially improves the performance of the student network by excluding features with larger distribution differences, but performance gradually declines as the number of features decreases. The \textit{False Mask} shows the worst performance, while the \textit{Random Mask} performs between the two.}
\label{fig:9}
\end{figure}

\begin{figure}[!htbp]
\centering
\includegraphics[width=0.95\linewidth]{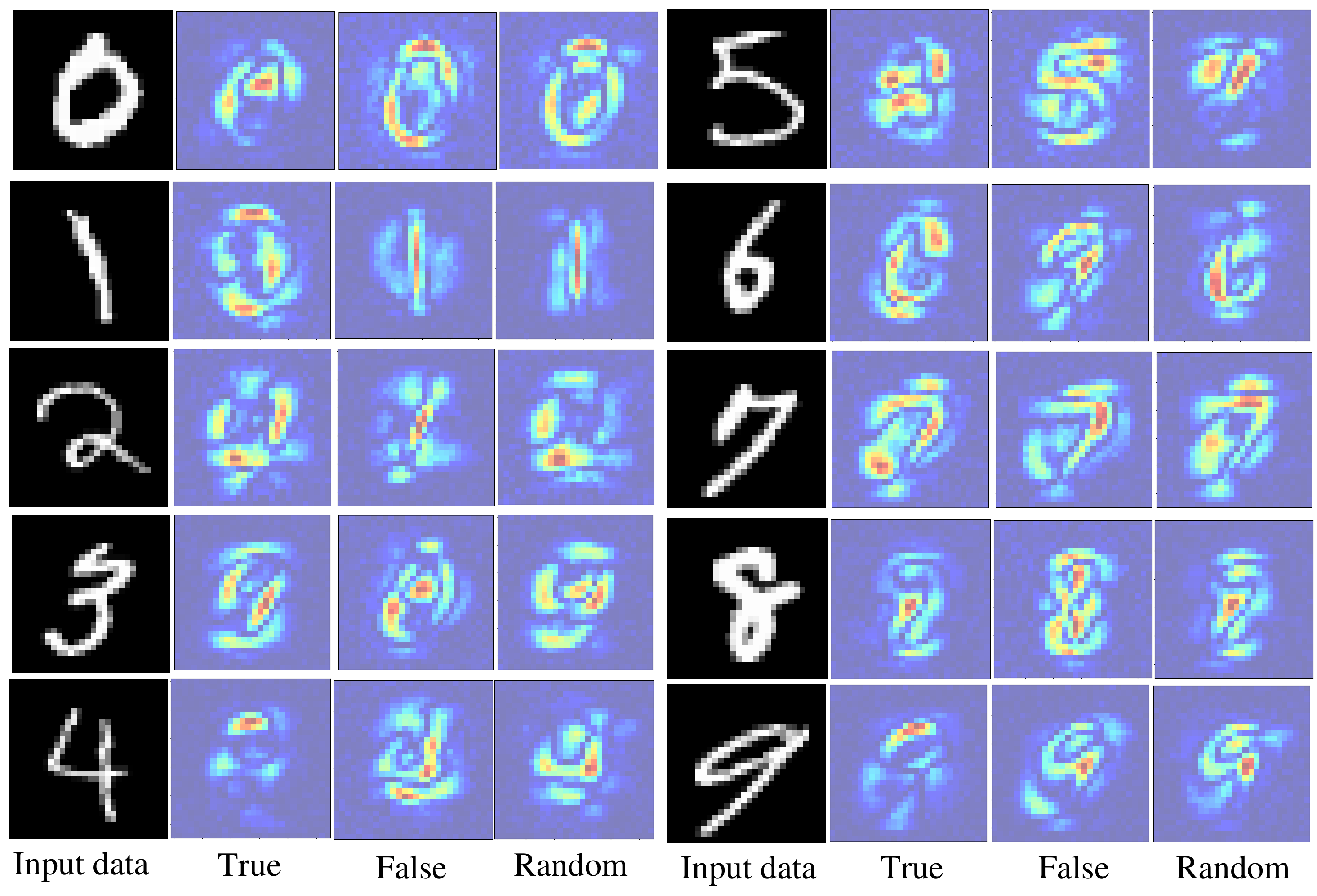}
\caption{The visualization results on the MNIST/MNIST-M datasets. Heatmaps highlight the masked feature regions. Specifically, the \textit{True Mask} represents feature regions with larger non-target distribution differences, the \textit{False Mask} represents regions with smaller differences, and the \textit{Random Mask} represents randomly selected feature regions.}
\label{fig:10}
\end{figure}

Similarly, we also conduct experiments on the MNIST/MNIST-M and RAVDESS datasets using Algorithm 1, and reach similar conclusions, as shown in Figs. \ref{fig:9}(a) and (b). Additionally, we use heatmaps to visualize the masked areas on the MNIST/MNIST-M datasets, as shown in Fig. \ref{fig:10}. For the \textit{False Mask}, the highlighted regions are mainly concentrated in the non-target areas, with fewer in the target areas; for the \textit{True Mask}, the situation is the opposite, with the highlighted regions primarily focused on the target areas and fewer in the non-target areas. From these results, it is observed that the target areas are mainly concentrated in the foreground, while the non-target areas are concentrated in the background, which aligns with our previous analysis.

\subsubsection{SemanticKITTI and NYU Depth V2}

\begin{figure}[!htbp]
\centering
\includegraphics[width=0.95\linewidth]{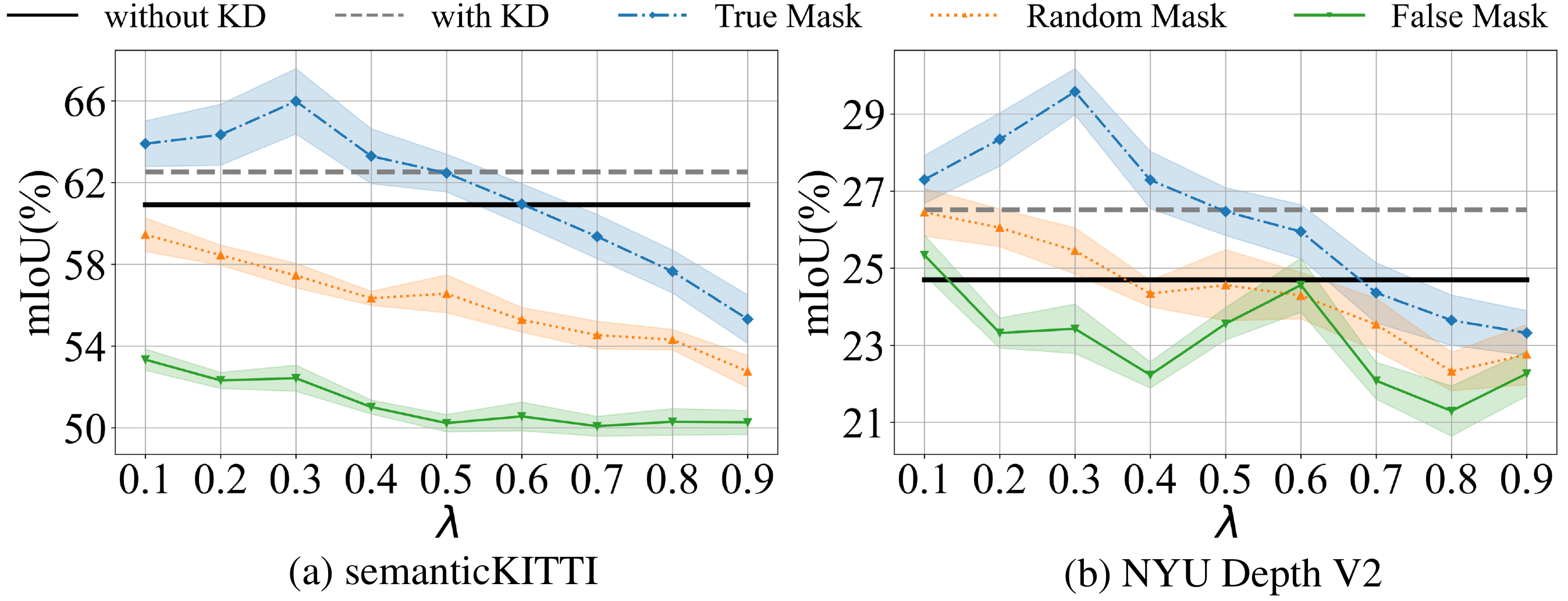}
\caption{Results of the sample masking method on the SemanticKITTI and NYU Depth V2 datasets. The \textit{True Mask} removes samples with larger non-target class distribution differences. The \textit{False Mask} and \textit{Random Mask} serve as control groups, representing the removal of samples with smaller non-target class distribution differences and the random removal of training samples, respectively.}
\label{fig:11}
\end{figure}

We use Algorithm 2 on the SemanticKITTI and NYU Depth V2 datasets, as shown in Figs. \ref{fig:11}(a) and (b). The results indicate that the effectiveness of KD initially improves and then declines as samples with large non-target class distribution divergence are removed. Initially, removing these samples enhances KD because their significant distribution differences hinder the process. However, as the removal proportion $\gamma\%$ increases, samples with smaller distribution divergences are also eliminated, leading to a decrease in KD effectiveness in later stages.

\begin{figure}[!htbp]
\centering
\includegraphics[width=0.95\linewidth]{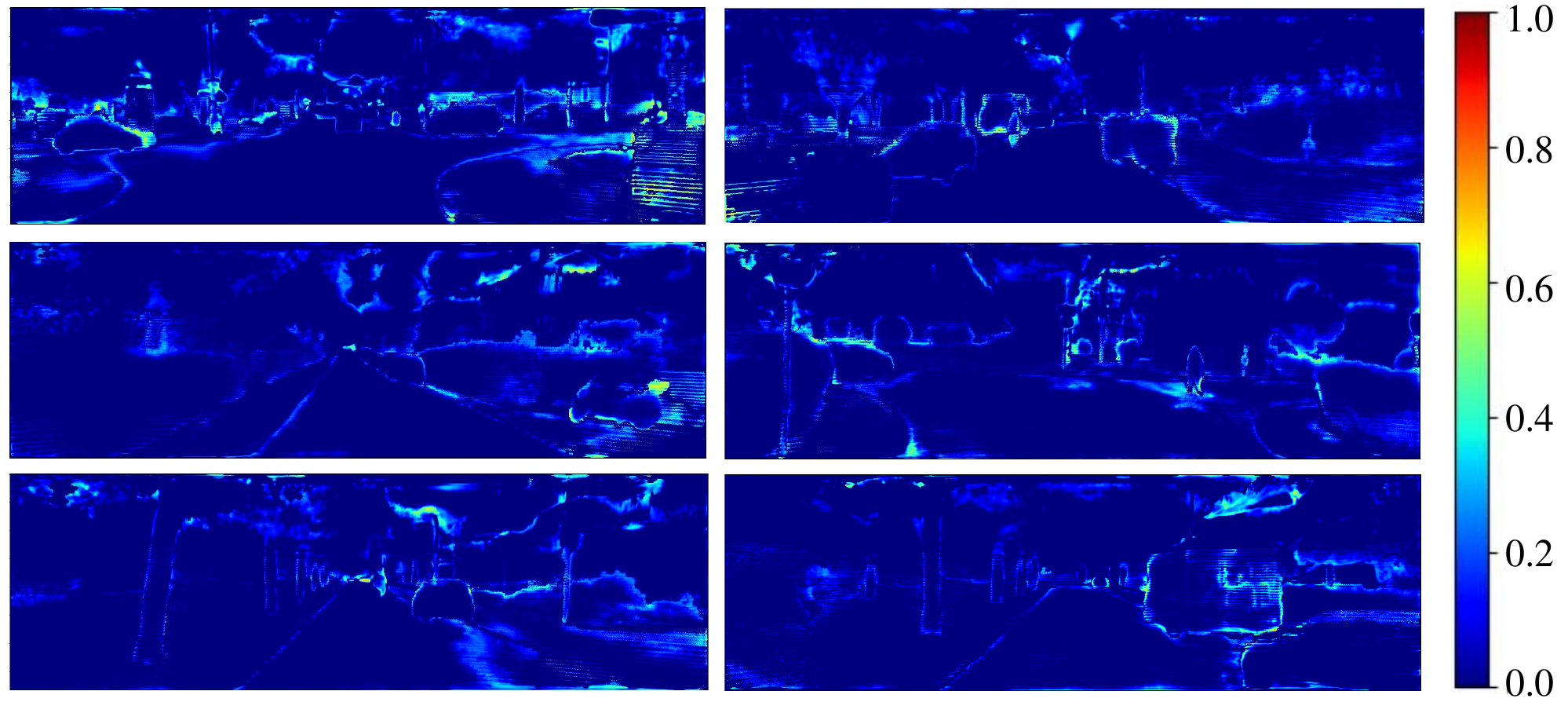}
\caption{The visualization results on the SemanticKITTI dataset. We highlight samples with significant differences in the distribution of non-target classes. The highlighted areas in the image are primarily concentrated at the edges of objects, indicating that the distribution differences between modalities are most pronounced in these edge regions.}
\label{fig:12}
\end{figure}

Furthermore, we visualize the samples with significant distribution divergence in non-target classes on the SemanticKITTI dataset, as shown in Fig. \ref{fig:12}. The highlighted areas, primarily concentrated along object edges, indicate that distribution divergence is greatest in these regions. The reason for this is that in segmentation tasks, the edges are typically the most challenging to distinguish, leading to the most pronounced divergence in non-target class distributions between different modalities.

\subsection{Generalization}

\par Beyond traditional KD \cite{hinton2015distilling}, various improved algorithms have been proposed. To evaluate the generalization ability of NTDH, we apply the feature masking method to several of these improved KD algorithms, including FitNet \cite{romero2014fitnets}, Contrastive Representation Distillation (CRD) \cite{tian2019contrastive}, Relational Knowledge Distillation (RKD) \cite{park2019relational}, Probabilistic Knowledge Transfer for deep representation learning (PKT) \cite{passalis2018probabilistic} , Similarity-Preserving KD (SP) \cite{tung2019similarity} and Decoupled Knowledge Distillation (DKD) \cite{zhao2022decoupled}.

\par Table \ref{tab:2} presents the experimental results on the MNIST/MNIST-M dataset. The results show that none of the methods improve student performance, indicating that KD methods designed for unimodal scenarios are not suitable for cross-modal applications. However, when we use Algorithm 1 to remove portions with significant non-target class distribution divergence, especially at smaller removal proportions (e.g., 25\% and 50\%), all algorithms show significant performance improvements. In contrast, random and false masking lead to substantial performance declines, highlighting the critical role of non-target class differences in cross-modal KD.

\par Table \ref{tab:3} presents the experimental results on the RAVDESS dataset. The results show that cross-modal KD fails to improve student performance, highlighting the limitations of these methods in cross-modal applications. However, using Algorithm 1 to remove portions with significant modal differences, particularly at lower removal proportions (e.g., 25\% and 50\%), leads to performance improvements in most algorithms. In contrast, control experiments with \textit{Random} and \textit{False} masking result in significant performance declines, underscoring the decisive role of non-target class distribution divergence in cross-modal KD.

\begin{table}[]
\centering
\caption{Experimental results on the MNIST/MNIST-M datasets. T, F, and R represent the true mask, false mask, and random mask, respectively.}
\label{tab:2}
\begin{tabular}{cc|c@{\hskip 8pt}c@{\hskip 8pt}c@{\hskip 8pt}c@{\hskip 8pt}c@{\hskip 8pt}l}
\toprule
\multicolumn{2}{c|}{} & FitNet  & CRD    & RKD     & PKT    & SP     & DKD    \\ \midrule
\multicolumn{2}{c|}{\centering w/o KD} & 69.1   & 69.1 & 69.1 & 69.1 & 69.1 & 69.1 \\
\multicolumn{2}{c|}{\centering w/o mask}  & 68.5  & 68.5  & 68.5  & 68.5  & 68.5  & 68.5  \\ \midrule
\multicolumn{1}{l|}{\multirow{3}{*}{\makecell{w mask \\ ($\lambda =0.25$)}}}  & T   & 70.3  & 71.3  & 70.1  & 59.85  & 72.4  & 75.8  \\
\multicolumn{1}{l|}{}   & R & 50.1  & 72.1  & 70.9  & 60.8  & 70.2  & 75.4  \\
\multicolumn{1}{l|}{}   & F  & 48.2  & 68.2  & 68.3  & 54.8  & 64.7  & 69.9  \\ \midrule

\multicolumn{1}{l|}{\multirow{3}{*}{\makecell{w mask \\ ($\lambda =0.50$)}}} & T   & 61.6  & 75.6  & 70.9  & 60.4  & 69.8  & 76.4  \\
\multicolumn{1}{l|}{}    & R & 50.1  & 70.1  & 71.4  & 56.3  & 68.5  & 73.2  \\
\multicolumn{1}{l|}{}  & F  & 43.6  & 63.6  & 69.3  & 53.5  & 62.7  & 69.6  \\ \midrule

\multicolumn{1}{l|}{\multirow{3}{*}{\makecell{w mask \\ ($\lambda =0.75$)}}} & T   & 59.3  & 70.3  & 70.1  & 57.1  & 65.6 & 72.7  \\
\multicolumn{1}{l|}{}    & R & 52.9  & 65.9 & 69.83  & 53.5  & 63.8  & 70.1  \\
\multicolumn{1}{l|}{}   & F  & 47.8  & 57.8  & 67.9  & 55.1  & 64.1  & 67.9  \\ \bottomrule
\end{tabular}
\end{table}

\begin{table}[]
\centering
\caption{Experimental results on the RAVDESS datasets. T, F, and R represent the true mask, false mask, and random mask, respectively.}
\label{tab:3}
\begin{tabular}{cc|c@{\hskip 8pt}c@{\hskip 8pt}c@{\hskip 8pt}c@{\hskip 8pt}c@{\hskip 8pt}l}
\toprule
\multicolumn{2}{c|}{} & FitNet  & CRD    & RKD    & PKT    & SP     & DKD    \\ \midrule
\multicolumn{2}{c|}{\centering w/o KD} & 71.4   & 71.4 & 71.4 & 71.4 & 71.4 & 71.4 \\
\multicolumn{2}{c|}{\centering w/o mask}  & 70.8  & 70.8  & 70.8  & 70.8  & 70.8  & 70.8  \\ \midrule

\multicolumn{1}{l|}{\multirow{3}{*}{\makecell{w mask \\ ($\lambda =0.25$)}}}  & T   & 81.7  & 75.3  & 69.8  & 80.3  & 79.6  & 77.8  \\
\multicolumn{1}{l|}{}   & R & 81.2  & 55.1  & 70.8  & 78.8  & 77.1  & 75.8  \\
\multicolumn{1}{l|}{}   & F  & 56.3  & 53.2  & 64.8  & 55.2  & 55.9  & 53.6  \\ \midrule

\multicolumn{1}{l|}{\multirow{3}{*}{\makecell{w mask \\ ($\lambda =0.50$)}}} & T   & 83.0  & 66.6  & 70.4  & 78.0  & 78.0  & 77.1  \\
\multicolumn{1}{l|}{}    & R & 77.7  & 55.1  & 66.3  & 78.0  & 76.8  & 78.2  \\
\multicolumn{1}{l|}{}  & F  & 57.5  & 48.6  & 63.5  & 55.4  & 57.1  & 54.8  \\ \midrule

\multicolumn{1}{l|}{\multirow{3}{*}{\makecell{w mask \\ ($\lambda =0.75$)}}} & T   & 63.5  & 64.3  & 67.1  & 62.9  & 63.6 & 72.3  \\
\multicolumn{1}{l|}{}    & R & 40.1  & 63.9  & 63.5  & 63.3  & 65.7  & 62.6  \\
\multicolumn{1}{l|}{}   & F  & 53.6  & 57.7  & 65.1  & 53.1  & 54.7  & 53.0  \\ \bottomrule
\end{tabular}
\end{table}

\subsection{Applications}

This section further explores the practical application of NTDH. According to this hypothesis, in cross-modal KD, if the non-target class distribution difference between Teacher (a) and the student is smaller than that between Teacher (b) and the student, then we expect the student guided by Teacher (a) to outperform the student guided by Teacher (b). To reduce the non-target class distribution difference between the teacher and student networks, the method in Algorithm 2 can be applied. To validate the effectiveness of this method, we apply it to existing cross-modal KD algorithms, such as 2DPASS \cite{yan20222dpass} and PMF \cite{zhuang2021perception}, and conduct tests on the SemanticKITTI dataset.

\begin{figure}[!htbp]
\centering
\includegraphics[width=0.95\linewidth]{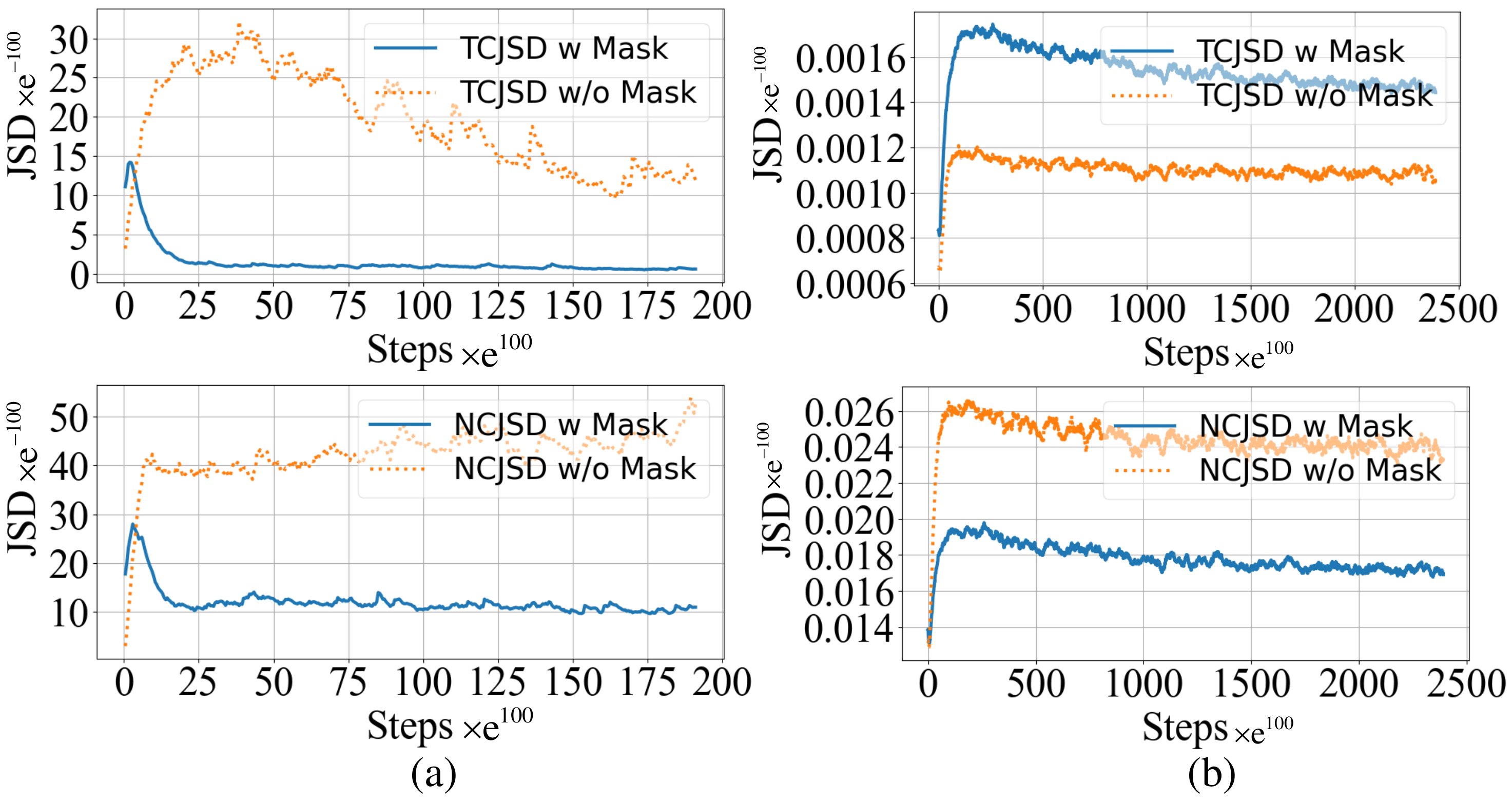}
\caption{Jensen-Shannon Divergence (JSD) between the teacher and student networks in both target and non-target distributions, denoted as TCJSD and NCJSD, respectively. (a) SemanticKITTI: After removing non-target samples with large distribution differences, NCJSD is significantly lower than TCJSD, indicating that the distribution differences among non-target classes have been greatly reduced. (b) NYU Depth V2: During the iterative process, the changes in NCJSD and TCJSD before and after applying the sample masking method are not significant, indicating that the original method has already removed non-target samples with large distribution differences.}
\label{fig:13}
\end{figure}

Table \ref{tab:4} shows the experimental results, indicating that appropriately masking samples with significant non-target distribution divergence (e.g., 25\% and 50\%) can improve performance. For 2DPASS, the maximum improvement reached 2.0\%. By comparing the change curves of NCJSD and TCJSD distances before and after masking, as shown in Fig. \ref{fig:13}(a), it is evident that the non-target distribution divergence significantly decreases, while the target distribution divergence changes only slightly. This confirms that the performance improvement is due to the reduction of non-target distribution differences. In contrast, the improvement in PMF is smaller because the Perception-Aware Loss in the original paper has already removed samples with large modality differences, leading to insignificant changes in NCJSD and TCJSD distances before and after masking, as shown in Fig. \ref{fig:13}(b).

\begin{table}[]
\centering
\caption{Application Experiment Results. We report the results of applying Algorithm 2 to the 2DPASS and PMF network models. The experiments show that removing appropriately selected non-target class samples effectively improves the performance of the original method. T, F, and R represent the true mask, false mask, and random mask, respectively.}
\label{tab:4}
\begin{tabular}{c|c|c@{\hskip 3pt}c@{\hskip 3pt}c|c@{\hskip 3pt}c@{\hskip 3pt}c|c@{\hskip 3pt}c@{\hskip 3pt}c}
\toprule
\multirow{2}{*}{} & \multirow{2}{*}{\makecell{Base \\ Line}} & \multicolumn{3}{c|}{\makecell{w mask \\ ($\lambda =0.25$)}} & \multicolumn{3}{c|}{\makecell{w mask \\ ($\lambda =0.50$)}} & \multicolumn{3}{c}{\makecell{w mask \\ ($\lambda =0.75$)}} \\ 
\cmidrule(lr){3-11} 
&   & T   & F   & R   & T   & F  & R  & T  & F  & R  \\ \midrule
2DPASS & 64.2 & 65.2   & 61.2    & 64.6     & 66.2   & 59.2   & 63.2    & 63.2  & 63.2   & 60.2    \\
PMF    & 63.2     & 64.2   & 60.2    & 57.2     & 65.2   & 58.2   & 62.2    & 62.2  & 60.2   & 63.2    \\ \bottomrule
\end{tabular}
\end{table}

\section{Conclusion}
In this work, we delve into cross-modal KD and introduce the NTDH method, which analyzes the impact of domain gaps in multimodal data on KD performance, highlighting the importance of non-target class distribution divergence. Through theoretical analysis and carefully designed experiments, we validate the rationale, generalization capability, and potential applications of NTDH. We aim for NTDH to provide valuable insights for the practical use of cross-modal KD and to foster interest in understanding domain discrepancies in multimodal learning. However, as this paper focuses on the effects of domain discrepancies on cross-modal KD, the exploration of theoretical applications is limited.

% \ifCLASSOPTIONcaptionsoff
%   \newpage
% \fi



%\bibliographystyle{IEEEtran}
\bibliographystyle{IEEEtran}
\bibliography{references}

\ifCLASSOPTIONcaptionsoff
  \newpage
\fi

\end{document}